\documentclass[10pt,twocolumn,letterpaper]{article}

\usepackage[pagenumbers]{wacv} 

\usepackage{graphicx}
\usepackage{amsmath}
\usepackage{amssymb}
\usepackage{booktabs}
\usepackage{algorithm}
\usepackage{algorithmic}
\usepackage{multirow} 
\usepackage{array}
\usepackage{tabularx}
\usepackage{float}
\usepackage{placeins}
\usepackage{stfloats}

\usepackage[pagebackref,breaklinks,colorlinks]{hyperref}
\usepackage{fancyhdr} 

\usepackage[capitalize]{cleveref}
\crefname{section}{Sec.}{Secs.}
\Crefname{section}{Section}{Sections}
\Crefname{table}{Table}{Tables}
\crefname{table}{Tab.}{Tabs.}

\fancypagestyle{firstpagestyle}{
    \fancyhf{}
    \fancyhead[C]{\Large\href{https://llvm-ad.github.io/}{Accepted to 2025 IEEE/CVF WACV LLVM-AD Workshop}}
}

\begin{document}

\title{Query3D: LLM-Powered Open-Vocabulary Scene Segmentation with Language Embedded 3D Gaussians}

\author{Amirhosein Chahe ~~~~~~ Lifeng Zhou\thanks{Corresponding author.}\\
Drexel University\\
Philadelphia PA 19104, USA\\
{\tt\small \{ac4462, lz457\}@drexel.edu}
}


\maketitle
\thispagestyle{firstpagestyle}

\begin{abstract}
This paper introduces a novel method for open-vocabulary 3D scene querying in autonomous driving by combining Language Embedded 3D Gaussians with Large Language Models (LLMs). We propose utilizing LLMs to generate both contextually canonical phrases and helping positive words for enhanced segmentation and scene interpretation. Our method leverages GPT-3.5 Turbo as an expert model to create a high-quality text dataset, which we then use to fine-tune smaller, more efficient LLMs for on-device deployment.
Our comprehensive evaluation on the WayveScenes101 dataset demonstrates that LLM-guided segmentation significantly outperforms traditional approaches based on predefined canonical phrases. Notably, our fine-tuned smaller models achieve performance comparable to larger expert models while maintaining faster inference times. Through ablation studies, we discover that the effectiveness of helping positive words correlates with model scale, with larger models better equipped to leverage additional semantic information.
This work represents a significant advancement towards more efficient, context-aware autonomous driving systems, effectively bridging 3D scene representation with high-level semantic querying while maintaining practical deployment considerations. Code and additional resources are available at \url{https://github.com/Zhourobotics/Query-3DGS-LLM}.
\end{abstract}
\section{Introduction}
Understanding complex scenes in three dimensions is crucial for advancements in autonomous systems, augmented reality, and robotics. The field of 3D scene understanding has seen significant advancements in recent years, particularly with the emergence of neural radiance fields (NeRF)~\cite{mildenhall2020nerfrepresentingscenesneural} and 3D Gaussian Splatting (3DGS)~\cite{kerbl20233dgaussiansplattingrealtime} for novel view synthesis. However, the challenge of open-vocabulary querying in 3D space remains critical for tasks such as object localization and segmentation in autonomous driving scenarios.

\begin{figure*}[h]
    \centering
        \includegraphics[width=1.0\linewidth, trim={0 2.1cm 0 3cm}, clip]{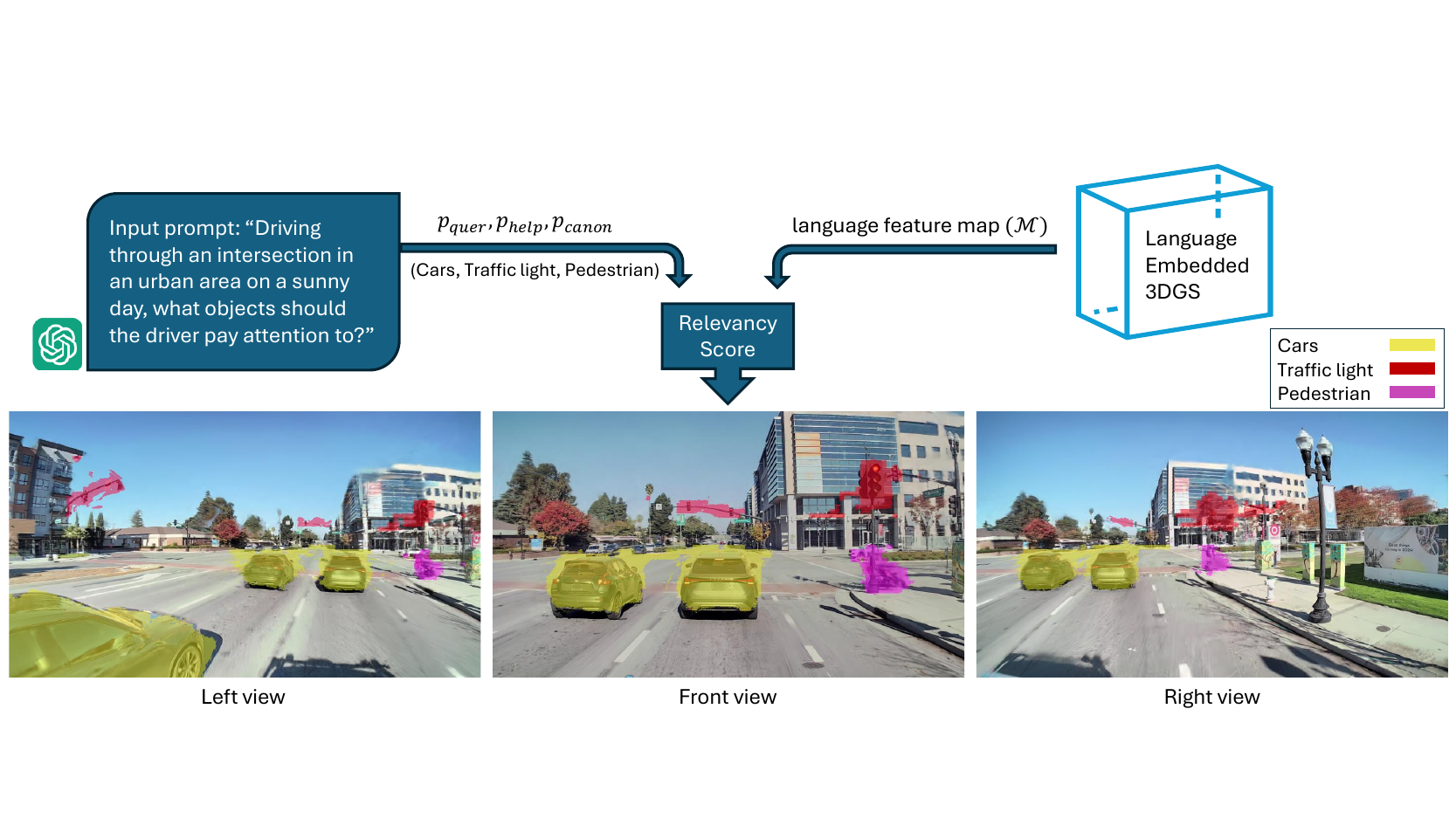}
    \caption{Overview of our LLM-enhanced Language Embedded 3D Gaussian Splatting pipeline. The LE3DGS model generates language feature maps, which are then processed using Algorithm \ref{alg:relevancy} to compute relevancy scores with LLM-generated queries ($p_{\texttt{quer}}$), helping positives ($p_{\texttt{help}}$), and canonical phrases ($p_{\texttt{canon}}$). The system responds to the high-level queries of a driving scenario by highlighting relevant objects (cars, traffic lights, pedestrians) across multiple views.}
    \label{fig:pipeline}
\end{figure*}

Concurrently, large language models (LLMs)~\cite{AttentionIA,PaLMSL,GPT3,lama2,mamba} have demonstrated remarkable abilities in grasping context and effectively analyzing and interpreting complex scenarios~\cite{LLM4DriveAS}. Recent works have begun to explore the integration of LLMs into autonomous driving tasks. For instance, NuPrompt~\cite{Wu2023LanguagePF}, a large-scale dataset for 3D object detection and tracking in autonomous driving, utilizes natural language prompts to predict object trajectories across different views and frames. While this approach outperforms traditional methods, it primarily uses LLMs to enhance 3D detection tasks rather than directly influencing tracking~\cite{LLM4DriveAS}. Further advancements in this direction include HiLM-D~\cite{Ding2023HiLMDTH}, which leverages multimodal LLMs (MLLMs)~\cite{zhu2023minigpt4enhancingvisionlanguageunderstanding,Jin2023ADAPTAD,Malla2022DRAMAJR} to improve autonomous driving tasks by integrating high-resolution information. Similarly, ToKEN~\cite{Tian2024TokenizeTW} introduces a novel MLLM that tokenizes the world into object-level knowledge, enabling better utilization of LLM's reasoning capabilities.

Recent works like Language Embedded Radiance Fields (LERF)~\cite{kerr2023lerflanguageembeddedradiance} and Language Embedded 3D Gaussians (LE3DGS)~\cite{shi2023le3dgaussians} have demonstrated the potential of language-embedded neural and 3D gaussians representations for scene understanding. Our study extends these approaches by introducing LLM-powered query processing to enhance open-vocabulary querying and segmentation for the scene understanding of autonomous vehicles. Our system supports both explicit queries that directly specify objects of interest (e.g., ``Show me the pedestrians'') and contextual queries that require scene understanding (e.g., ``What should I watch for at this intersection?''). For either type of query, our approach employs LLMs to dynamically generate three complementary types of semantic references: query words that capture the core search intent, context-aware canonical phrases that provide semantic anchors, and helping positive words that enhance the segmentation further. Figure~\ref{fig:pipeline} demonstrates our approach processing a contextual query. The LLM first interprets the user's query to identify relevant scene elements, then generates appropriate canonical phrases and helping positive words. This combination enables more precise object segmentation than traditional fixed-phrase methods, particularly for contextual queries that require deeper scene understanding.

By integrating the contextual reasoning capabilities of LLMs with the spatial precision of 3D Gaussians, our method significantly improves the querying capabilities in driving scenes. This integration enables more flexible and context-aware scene interpretation while providing an efficient path toward practical deployment. The main contributions of our work are as follows:

\begin{itemize}
    \item We enhance LE3DGS with LLM-generated language guidance. Unlike previous approaches that use fixed canonical phrases, our LLM dynamically generates two types of descriptive language: canonical phrases to differentiate scene elements and helping positive words for additional context. These LLM-generated phrases are more contextually suited for LE3DGS's language embeddings, leading to more accurate scene segmentation. This is particularly effective for complex queries where traditional fixed phrases might miss subtle distinctions.
    \item We develop an effective fine-tuning strategy that enables smaller LLMs to achieve performance comparable to larger expert models while maintaining faster inference times for on-device deployment in autonomous vehicles.
    
    \item Through comprehensive experiments on the WayveScene101 dataset, we demonstrate that our LLM-guided approach significantly outperforms traditional fixed canonical phrase methods. Our ablation studies reveal that the effectiveness of helping positive words correlates with model scale, with larger models better equipped to leverage additional semantic context.
\end{itemize}

\section{Related Work}

\subsection{3D Scene Representations with Semantic Features}
Recent advances in 3D scene representation have focused on incorporating semantic information alongside geometry and appearance. Neural radiance fields were extended to encode semantics~\cite{zhi2021inplacescenelabellingunderstanding}, enabling 2D semantic labels with minimal annotations, while Neural Feature Fusion Fields (N3F)~\cite{tschernezki2022neuralfeaturefusionfields} improved feature extraction through 2D-to-3D distillation. 3D Gaussian Splatting~\cite{kerbl20233dgaussiansplattingrealtime} emerged as an efficient alternative to NeRF, offering comparable quality with faster processing~\cite{malarz2024gaussiansplattingnerfbasedcolor}.

The integration of language features into these representations marked a significant advancement. Language Embedded Radiance Fields (LERF)~\cite{kerr2023lerflanguageembeddedradiance} pioneered open-ended language queries by grounding CLIP~\cite{radford2021learning} embeddings into NeRF. Feature 3DGS~\cite{zhou2024feature3dgssupercharging3d} and Language Embedded 3D Gaussians (LE3DGS)~\cite{shi2023le3dgaussians} extended this concept to Gaussian splatting, enabling efficient language-guided editing and semantic segmentation. ConceptFusion~\cite{jatavallabhula2023conceptfusionopensetmultimodal3d} and FeatureNeRF~\cite{Ye_2023} further advanced cross-modal reasoning by leveraging foundation models.

\subsection{LLM-based 3D Object Grounding}
Recent works have explored LLMs for 3D object grounding tasks with different approaches~\cite{Guo2023ViewReferGT,Bakr2023CoT3DRefCD,Jia2024SceneVerseS3,Yin2023LAMMLM}. ViL3DRel~\cite{vil3drel} incorporates spatial relationships through a transformer-based architecture with spatial self-attention. LLM-Grounder~\cite{llmgrounder} and VLM-Grounder~\cite{vlmgrounder} employ LLMs as reasoning agents for grounding, with the former decomposing complex queries into subtasks and the latter focusing on efficient multi-view processing. Object-centric approaches~\cite{3dsceneobjectlevel,opensceneunderstanding} represent scenes through distinct object entities with unique identifiers, combining 2D and 3D features for comprehensive scene understanding. While these methods excel at grounding specific objects based on language descriptions, our work differs by focusing on enhancing open-vocabulary querying capabilities through LLM-generated canonical phrases and helping positives, enabling more flexible scene interpretation for autonomous driving applications.

\subsection{Limitations of Current Approaches}
Despite these advances, current methods including LE3DGS~\cite{shi2023le3dgaussians}, LERF~\cite{kerr2023lerflanguageembeddedradiance}, and LangSplat~\cite{qin2024langsplat3dlanguagegaussian} share several key limitations. They rely on fixed predefined canonical phrases during inference, which limits their adaptability to both explicit and contextual queries. Their underlying vision-language models exhibit ``bag-of-words'' behavior~\cite{bagofthewords}, treating queries as collections of independent words rather than understanding their semantic relationships. Additionally, the reliance on CLIP's text encoder, with its effective usable length of just 20 tokens~\cite{zhang2024longclipunlockinglongtextcapability}, restricts these methods to very short queries.

Our work addresses these limitations by leveraging LLMs to dynamically generate canonical phrases and helping positive words at inference time. By integrating LLMs with LE3DGS, we enable a system that can effectively process both explicit and contextual queries while adapting its language understanding to specific scene requirements, capturing subtle details that the systems with fixed canonical phrases might miss.

\begin{figure*}
    \centering

\setlength{\lineskip}{4pt}
\includegraphics[width=0.26\linewidth]{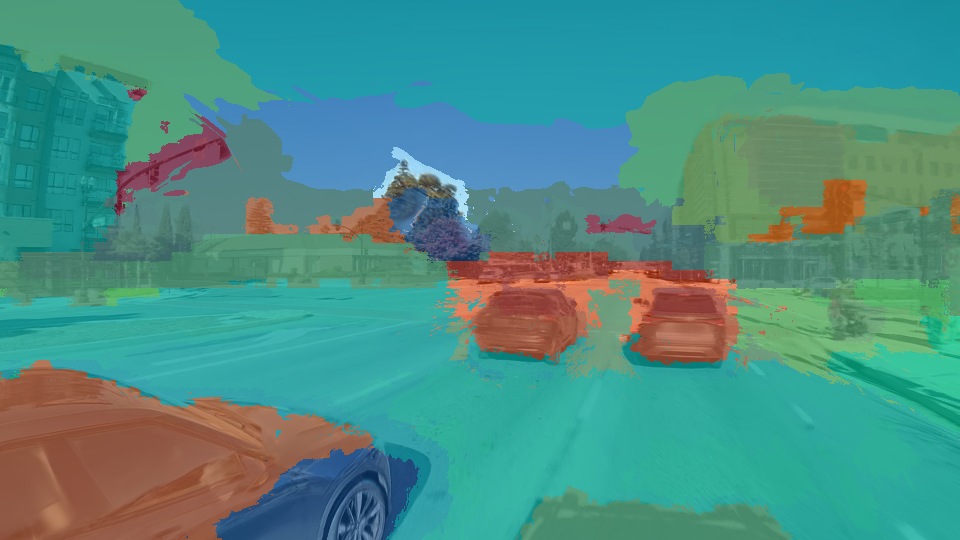}\hspace{0.002\linewidth}
\includegraphics[width=0.26\linewidth]{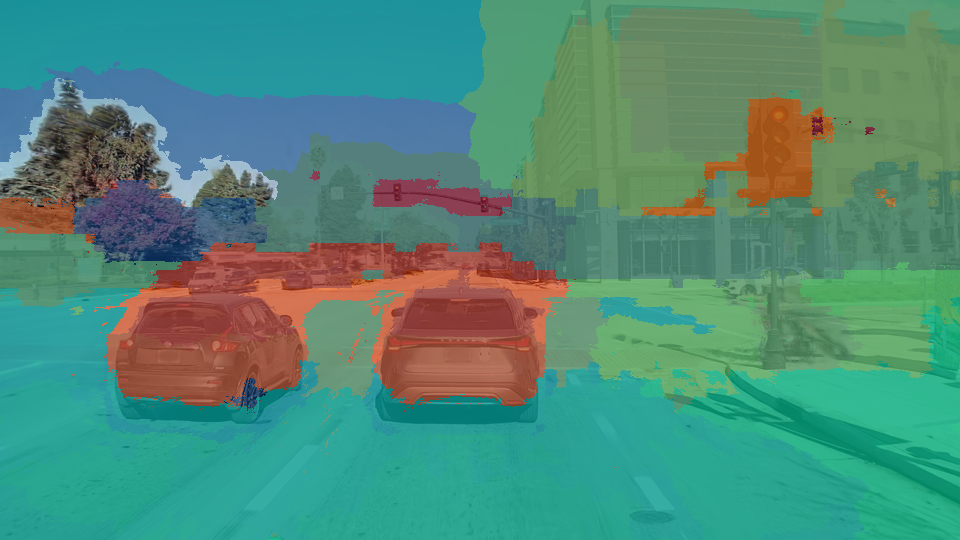}\hspace{0.002\linewidth}
\includegraphics[width=0.26\linewidth]{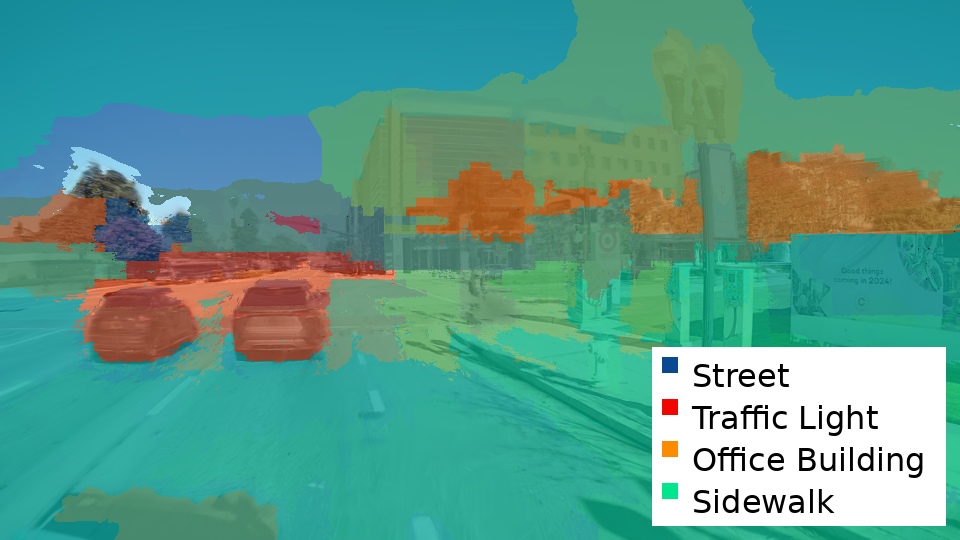}
\\
\includegraphics[width=0.26\linewidth]{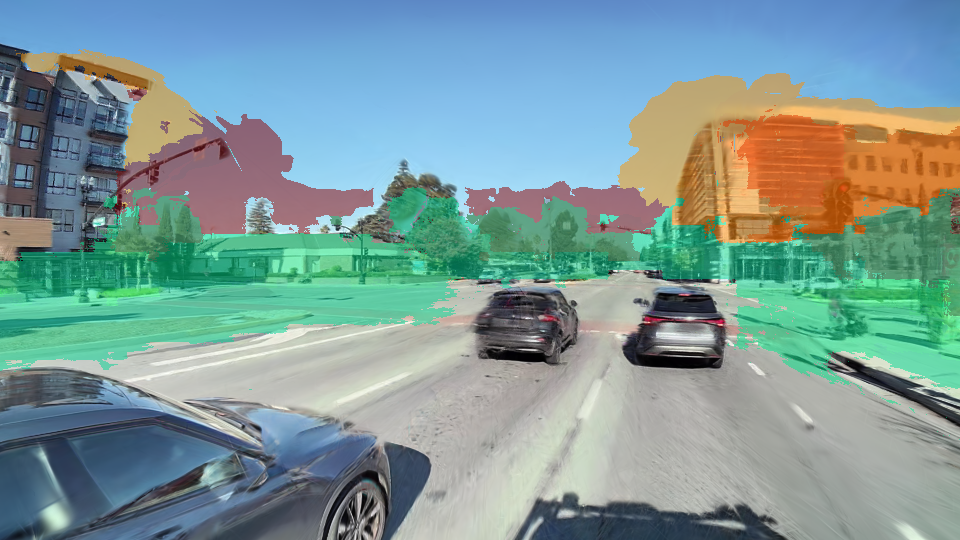}\hspace{0.002\linewidth}
\includegraphics[width=0.26\linewidth]{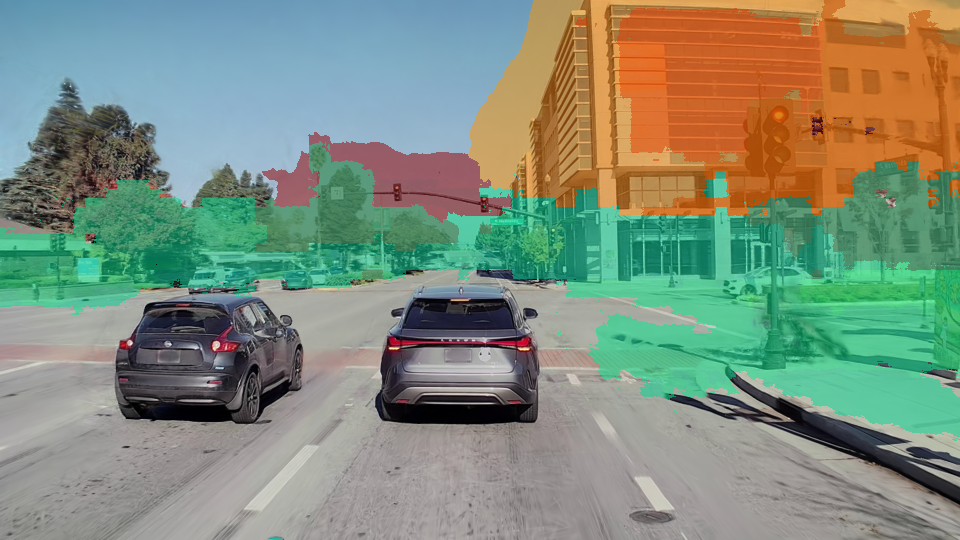}\hspace{0.002\linewidth}
\includegraphics[width=0.26\linewidth]{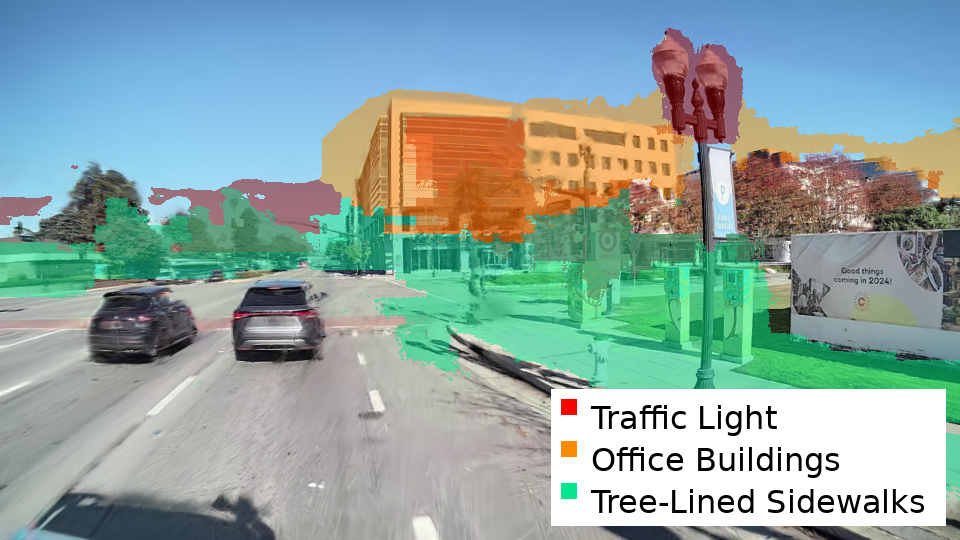}
\\
\includegraphics[width=0.26\linewidth]{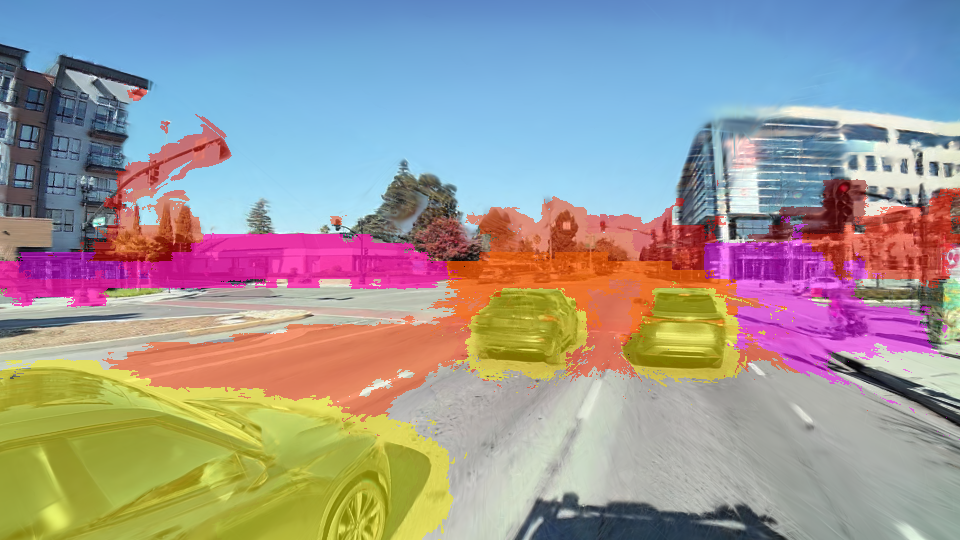}\hspace{0.002\linewidth}
\includegraphics[width=0.26\linewidth]{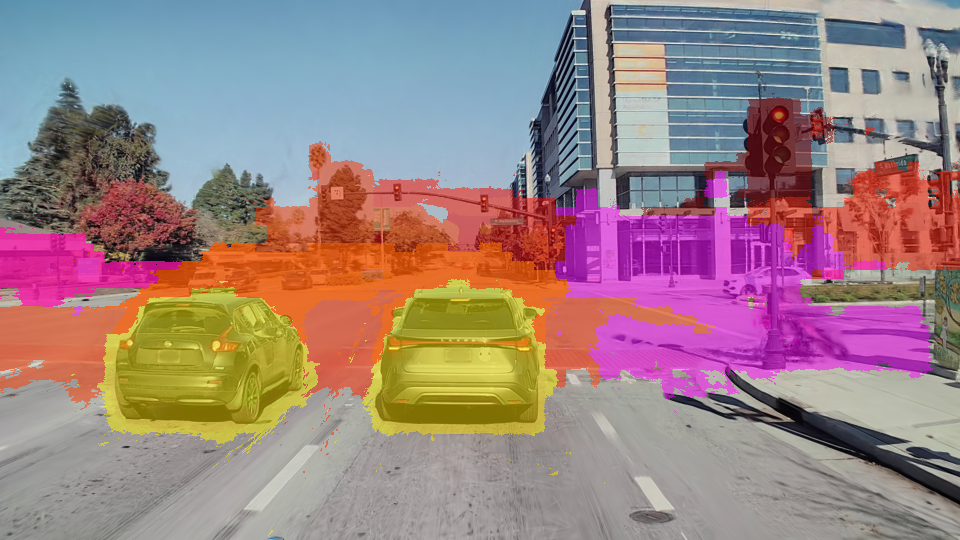}\hspace{0.002\linewidth}
\includegraphics[width=0.26\linewidth]{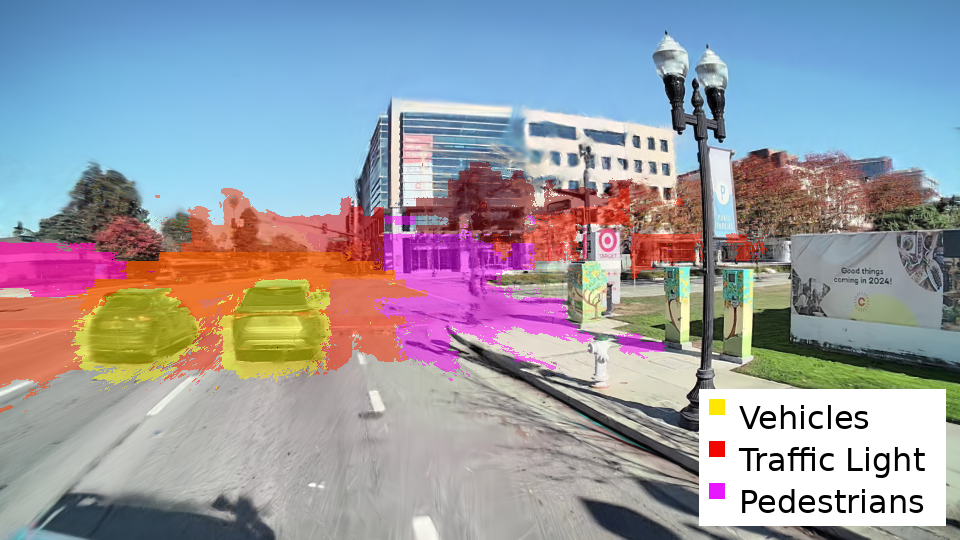}
\\
\includegraphics[width=0.26\linewidth]{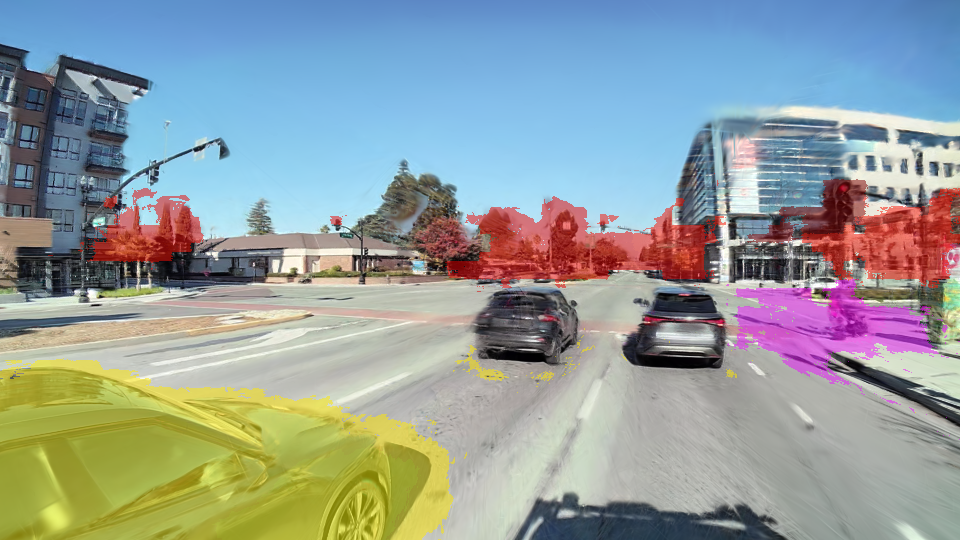}\hspace{0.002\linewidth}
\includegraphics[width=0.26\linewidth]{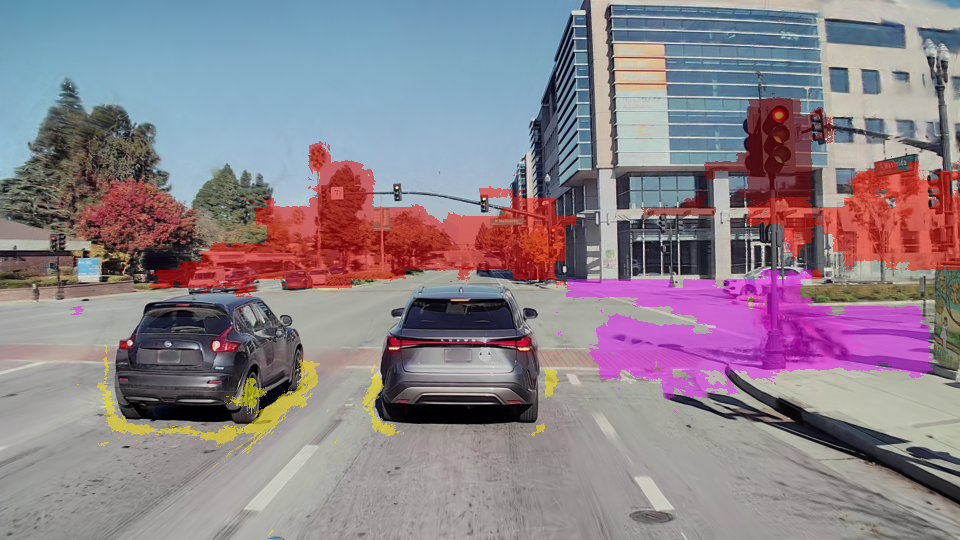}\hspace{0.002\linewidth}
\includegraphics[width=0.26\linewidth]{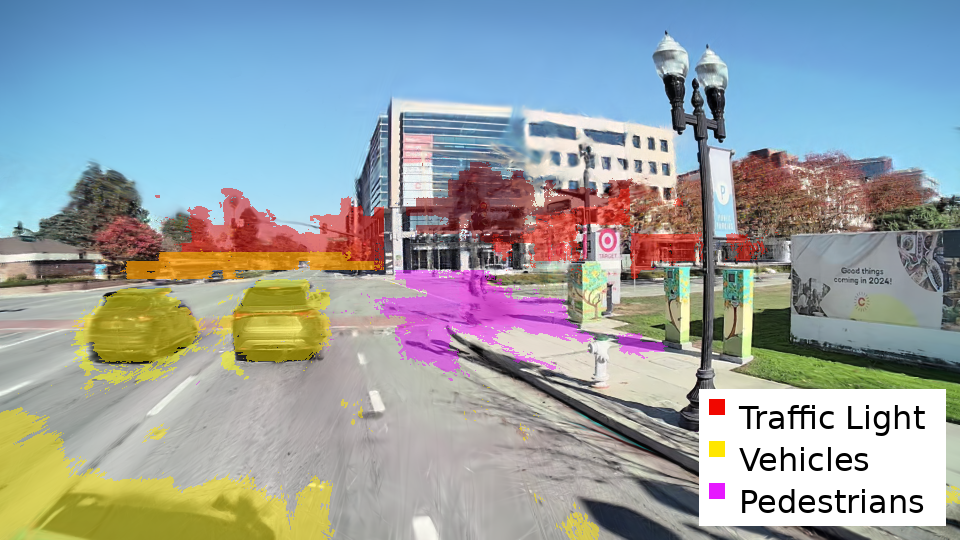}

\caption{Visual comparison of segmentation results using different Qwen models responding to a contextual query ``Driving through an intersection in an urban area on a sunny day, what objects should the driver pay attention to?'' From top to bottom: Results from Qwen-0.5B, 1.5B, 3B, and 7B models. Each row shows three camera views (left, front, right) of the same scene. The results demonstrate how larger models produce more precise and contextually relevant segmentations for autonomous driving scenarios.}

\end{figure*}

\section{Methodology}
\subsection{Scene Representation with 3D Gaussians}
While our method is compatible with various neural representations like LERF~\cite{kerr2023lerflanguageembeddedradiance}, we leverage LE3DGS~\cite{shi2023le3dgaussians} for its efficient inference capabilities to generate 3D Gaussians with language features integrated. This method represents the scene using language-embedded Gaussian points and allows for both open-vocabulary querying and high-quality novel view synthesis in an efficient manner.
The method consists of several key components as follows:
\subsubsection{3D Gaussian Splatting}
The method builds upon the efficient 3D Gaussian Splatting technique~\cite{kerbl20233dgaussiansplattingrealtime} for high-quality novel view synthesis, which represents the scene with 3D Gaussians initialized from a sparse point cloud generated via Structure-from-Motion (SfM). Each Gaussian is characterized by its position, anisotropic covariance, opacity, and color via spherical harmonics. Such a representation gives rise to an efficiently differentiable projection to 2D for real-time rendering. The final image derives from the alpha blending model used in volumetric rendering, represented as:
\begin{equation}
C = \sum_{i=1}^{N} T_i (1 - \exp(-\sigma_i \delta_i)) c_i
\end{equation}
where $C$ is the accumulated color along the viewing ray, $T_i$ is the accumulated transmittance to the $i$-th sample. $\sigma_i$ and $\delta_i$ denote the density and segment length of the $i$-th Gaussian, and $c_i$ is the color contribution of the $i$-th Gaussian.

\subsubsection{Dense Language Feature Extraction}
Dense language features are extracted from multi-view images using both CLIP~\cite{radford2021learning} and DINO~\cite{caron2021emerging} models. These features can provide rich semantic information about the scene.

\subsubsection{Feature Quantization}
To reduce the high memory requirement of the raw language feature embeddings on dense 3D Gaussians, a quantization scheme~\cite{oord2018neuraldiscreterepresentationlearning} is proposed. It creates a discrete feature space, yielding semantic information compression without losing its effectiveness.

\subsubsection{Compact Semantic Features on 3D Gaussians}
Instead of directly embedding high-dimensional language features, LE3DGS learns compact semantic feature vectors for each 3D Gaussian. These are then decoded into discrete semantic indices using a small MLP.

\subsubsection{Adaptive Spatial Smoothing}
To handle semantic ambiguities arising from multi-view inconsistencies, LE3DGS introduces an adaptive smoothing mechanism~\cite{shi2023le3dgaussians}. The smoothing loss is:
\begin{equation}
    L_{smo} = \|s_{MLP} - s^*_G\|^2 + \max(u^*_G, w_s)\|s^*_{MLP} - s_G\|^2
\end{equation}
where $s_{MLP}$ are smoothed semantic features generated by an MLP, $s_G$ represents the original semantic features of the 3D Gaussians, $u_G$ is the learned uncertainty value for each 3D Gaussian, and $w_s$ is a minimal weight for the semantic smoothing term. The $*$ operator denotes stop gradient, preventing backpropagation through these terms.

\subsubsection{Optimization}
It performs a joint optimization over 3D Gaussians for appearance and semantic information. The total loss of the reconstruction and the semantics is:
\begin{equation}
    \mathcal{L} = \lambda_s \mathcal{L}_s + \lambda_{smo} \mathcal{L}_{smo}.
\end{equation}
$\mathcal{L}_s$ is the semantic loss:
\begin{equation}
\mathcal{L}_s = \lambda_{CE} \mathcal{L}_{CE} + \lambda_u \mathcal{L}_u
\end{equation}
where $\mathcal{L}_{CE}$ is cross entropy loss and $\mathcal{L}_u$ is the uncertainty regularization loss. For further details, the reader is referred to~\cite{shi2023le3dgaussians}.
\subsection{Inference}
During inference, we compute a language feature map $\mathcal{F}$ from the scene representation:
\begin{equation}
\mathcal{F}=\mathcal{M}_{\text {infer }} \mathbf{S}
\end{equation}
where $\mathcal{M}_{\text {infer }}$ is the inferred distribution of semantic indices and $\mathbf{S}$ is the matrix of quantized language features.

The relevancy score for scene elements is calculated following~\cite{kerr2023lerflanguageembeddedradiance}:
\begin{equation}
\text{relevancy} = \min_{i}\frac{\exp(\mathcal{F} \cdot \phi_{\texttt{quer}})}{\exp(\mathcal{F} \cdot \phi^i_{\texttt{canon}}) + \exp(\mathcal{F} \cdot \phi_{\texttt{quer}})}
\label{eq:rele}
\end{equation}
where $\phi_{\texttt{quer}}$ is the CLIP embedding of the query text, $\phi^i_{\texttt{canon}}$ are embeddings of canonical phrases, and $\cdot$ represents cosine similarity.

\subsection{LLM-Enhanced Open-Vocabulary Querying}
Previous approaches~\cite{shi2023le3dgaussians,kerr2023lerflanguageembeddedradiance,qin2024langsplat3dlanguagegaussian} rely on fixed canonical phrases (``object", ``things", ``stuff", ``texture"), limiting their ability to handle diverse scenarios. Additionally, these methods often exhibit the ``bag-of-words" behavior~\cite{llmgrounder}, treating queries as collections of independent words rather than understanding their semantic relationships.

Our system processes both explicit queries that directly specify objects (e.g., ``Show me the traffic signs'') and contextual queries that require scene interpretation (e.g., ``What should I watch for in this construction zone?''). For either query type, our LLM generates:

\begin{itemize}
   \item Canonical phrases for computing relevancy scores to differentiate scene elements.
   \item Helping positive words that enhance matching with LE3DGS's language embeddings.
\end{itemize}

For example, when processing either an explicit query ``Show me the traffic signs'' or interpreting a contextual query about a construction zone, our LLM generates:
\begin{itemize}
   \item Canonical phrases to differentiate elements: ``construction cones", ``traffic cones", ``pedestrians", ``vehicles", ``traffic lights", and
   \item Helping positive words for better matching: ``road signs", ``warning signs", ``directional markers".
\end{itemize}

The rationale for this approach is twofold. First, unlike CLIP's limited token capacity~\cite{zhang2024longclipunlockinglongtextcapability}, LLMs can process longer, more nuanced descriptions to generate appropriate phrases. Second, LLMs understand semantic relationships between concepts, addressing the bag-of-words limitation~\cite{bagofthewords,kerr2023lerflanguageembeddedradiance} by providing related terms that help distinguish fine-grained differences. This LLM-driven approach enables more accurate segmentation by adapting to both explicit and contextual queries while maintaining precise object differentiation through canonical phrases and helping positive words. Its detailed steps are shown in Algorithm~\ref{alg:relevancy}

\begin{algorithm}[h]
\caption{Query Processing and Relevancy Score Calculation}
\label{alg:relevancy}
\begin{algorithmic}[1]
\REQUIRE {language feature map: $\mathcal{F}$, user query (explicit or contextual)}
\ENSURE{relevancy score}
\STATE $p_{\texttt{quer}}, p_{\texttt{help-pos}}, p_{\texttt{canon}} \gets \text{LLM}(\text{query})$ 
\label{alg:line: llm}
\STATE $\phi_{\texttt{quer}} \gets \text{CLIP}_{\text{text encoder}}(p_{\texttt{quer}})$
\label{alg:line: clip1}
\STATE $\phi_{\texttt{help-pos}} \gets \text{CLIP}_{\text{text encoder}}(p_{\texttt{help-pos}})$
\label{alg:line: clip2}
\STATE $\phi_{\texttt{canon}} \gets \text{CLIP}_{\text{text encoder}}(p_{\texttt{canon}})$
\label{alg:line: clip3}
\STATE $\phi_{\texttt{pos}} \gets \text{CONCATENATE}(\phi_{\texttt{quer}}, \phi_{\texttt{help-pos}})$
\label{alg:line: concat}
\STATE $\sigma_{\texttt{pos}} \gets \text{COSINE SIMILARITY}(\phi_{\texttt{pos}}, \mathcal{F})$
\label{alg:line: cos1}
\STATE $\sigma_{\texttt{canon}} \gets \text{COSINE SIMILARITY}(\phi_{\texttt{canon}}, \mathcal{F})$
\label{alg:line: cos2}
\STATE $\text{score} \gets  \min_{i}\left(\frac{\exp(\sigma_{\texttt{pos}})}{\exp(\sigma_{\texttt{canon}}^{i}) + \exp(\sigma_{\texttt{pos}})}\right)$
\label{alg:line: score}
\RETURN score
\end{algorithmic}
\end{algorithm}

Algorithm~\ref{alg:relevancy} processes user queries and calculates relevancy scores for scene elements. The algorithm accepts both explicit queries (e.g., ``Show me pedestrians'') and contextual queries (e.g., ``What should I watch at this intersection?''). In Line~\ref{alg:line: llm}, the LLM processes the user query to generate three outputs: primary query $p_{\texttt{quer}}$ identifying objects of interest, helping positive words $p_{\texttt{help-pos}}$ for enhanced matching, and canonical phrases $p_{\texttt{canon}}$ for differentiation.
In Lines~\ref{alg:line: clip1}--\ref{alg:line: clip3}, these phrases are encoded using CLIP's text encoder to obtain their respective embeddings. Line~\ref{alg:line: concat} concatenates the primary query and helping positive embeddings into $\phi_{\texttt{pos}}$ to enhance matching accuracy. Lines~\ref{alg:line: cos1}--\ref{alg:line: cos2} compute cosine similarities between the language feature map and both the combined positive embeddings ($\sigma_{\texttt{pos}}$) and canonical embeddings ($\sigma_{\texttt{canon}}$).
Finally, Line~\ref{alg:line: score} applies the softmax-like formula shown in Eq.~\ref{eq:rele} to calculate the relevancy score, representing the probability of each pixel corresponding to the queried objects. The resulting scores can be thresholded to obtain the final segmentation.

\begin{figure}[h]
\centering
\fbox{
\begin{minipage}{0.95\columnwidth}
\vspace{0.5em}
\textbf{System:} Object detection assistant

\textbf{Scene:} ``Driving through an urban street with construction work, traffic cones, and barriers on an overcast day. The road has temporary lane markings and directional signs, with buildings on both sides and leafless trees along the street.''

\textbf{Task:} For a given object:
\begin{itemize}
    \item Provide up to 3 helping positives, and
    \item Provide up to 6 Canonical phrases.
\end{itemize}

\textbf{Query:} Object: “Pedestrian''

\textbf{Response:} 
\texttt{Helping positives: person, human, crosswalk}\\
\texttt{Canonical phrases: traffic light, road, traffic cone, building, tree}
\vspace{0.5em}
\end{minipage}
}
\caption{Example conversation with GPT-3.5 Turbo to query ``Pedestrian'' from scene 12 in WayveScene101.}
\label{fig:gpt_conversation}
\end{figure}

\begin{table*}[b]
\small
\centering
\caption{Performance comparison of different model variants with and without helping positives. The table shows metrics for both instruction-tuned and fine-tuned Qwen and Llama models of varying sizes (0.5B-7B parameters) and GPT-3.5 Turbo against baseline segmentation. The best values in each column are in \textbf{bold}, second-best values are \underline{underlined}.}
\begin{tabular}{l|cccc|cccc}
\hline
\multirow{2}{*}{Model} & \multicolumn{4}{c|}{Without Helping Positives} & \multicolumn{3}{c}{With Helping Positives} \\
\cline{2-8}
& IoU & Accuracy & Precision & mAP & IoU & Accuracy & Precision & mAP \\
\hline
Qwen2.5-0.5B-Instruct & 0.16 & 0.68 & 0.22 & 0.42 & 0.08 & 0.75 & 0.13 & 0.32\\
Qwen2.5-1.5B-Instruct & 0.19 & 0.78 & 0.23 & 0.46 & 0.07 & 0.88 & 0.10 & 0.38\\
Qwen2.5-3B-Instruct & 0.18 & 0.75 & 0.24 & 0.47 & 0.05 & 0.77 & 0.15 & 0.34\\
Qwen2.5-7B-Instruct & 0.25 & 0.83 & 0.32 & 0.53 & 0.15 & 0.88 & 0.24 & 0.47\\
\hline
Qwen2.5-0.5B-Finetuned & 0.17 & 0.86 & 0.24 & 0.51 & 0.15 & 0.87 & 0.22 & 0.42\\
Qwen2.5-1.5B-Finetuned & \textbf{0.28} & 0.81 & \textbf{0.37} & \underline{0.54} & 0.17 & 0.81 & 0.28 & 0.43\\
Qwen2.5-3B-Finetuned & 0.27 & 0.86 & 0.35 & \textbf{0.56} & 0.17 & 0.86 & 0.24 & 0.45\\
Qwen2.5-7B-Finetuned & 0.24 & \underline{0.86} & 0.33 & 0.53 & \underline{0.23} & \underline{0.89} & \underline{0.34} & \underline{0.47} \\
\hline
Llama3.2-1B-Instruct & 0.18 & 0.71 & 0.20 & 0.46 & 0.16 & 0.73 & 0.19 & 0.42\\
Llama3.2-3B-Instruct & 0.21 & 0.79 & 0.27 & 0.51 & 0.12 & 0.86 & 0.24 & 0.41\\
\hline
Llama3.2-1B-Finetuned& 0.18 & 0.76 & 0.24 & 0.49 & 0.12 & 0.81 & 0.22 & 0.40\\
Llama3.2-3B-Finetuned& 0.26 & 0.84 & 0.32 & 0.53 & 0.18 & 0.87 & 0.26 & 0.43\\
\hline
GPT-3.5 Turbo & \underline{0.27} & \textbf{0.90} & \underline{0.37} & 0.54 & \textbf{0.32} & \textbf{0.94} & \textbf{0.50} & \textbf{0.54} \\
\hline
Base & 0.18 & 0.64 & 0.19 & 0.50 & 0.18 & 0.64 & 0.19 & 0.50 \\
\hline
\end{tabular}
\label{tab:performance}
\end{table*}

\section{Experiments}
\subsection{Dataset}
\subsubsection{WayveScene}
We use a subset of the WayveScenes101 dataset \cite{zürn2024wayvescenes101datasetbenchmarknovel}, designed specifically to advance novel view synthesis in challenging real-world driving scenarios. We selected a subset of five representative scenes (IDs: 12, 21, 36, 78 and 88) totaling about 500 images. This selection balances computational feasibility with the need for diverse and challenging data, encompassing various environmental conditions, lighting variations, and complex urban environments. This allows us to thoroughly evaluate our approach in handling the complexities of autonomous driving environments. 
For quantitative evaluation of our method's semantic understanding capabilities, we generate segmentation maps for the images in the evaluation set using a pre-trained Segformer model \cite{xie2021segformer} on the Cityscapes dataset \cite{cordts2016cityscapesdatasetsemanticurban}. This semantic information allows us to assess the semantic accuracy of our proposed method.
\subsubsection{Text Dataset}
We generate a set of helping positives and canonical phrases for each query object 
using GPT-3.5 Turbo~\cite{openai2024gpt35} for fine-tuning. For each primary query, the model provided 1 to 3 helping positive words and up to 6 canonical phrases. Figure \ref{fig:gpt_conversation} denotes the structure of one example for scene 12 from WayveScene101 in the generated dataset. The ``Scene'' description could be provided by the user or using an image captioning model.

\subsection{LLMs}
In addition to GPT-3.5 Turbo, we employ a variety of smaller, open-source models to improve inference efficiency and support on-device computation for applications such as autonomous vehicles and robotics. Specifically, we utilize instruction-tuned Qwen2.5~\cite{qwen2,qwen2.5}, available in parameter sizes of 0.5, 1.5, 3, and 7 billion, along with instruction-tuned Llama 3.2~\cite{lama3herdmodels}, offered in 1 and 3 billion parameter configurations. To optimize performance further, we fine-tune these models on a custom text dataset that we generated from GPT-3.5 Turbo.

\subsubsection{Fine-tuning Process}
We fine-tune the instruction-tuned Qwen and Llama models using LoRA~\cite{lora} (Low-Rank Adaptation) for efficient parameter updates. We leverage Unsloth~\cite{unsloth}, which optimizes LoRA's backward propagation, reducing FLOPs during gradient descent. The key projection layers (\texttt{q\_proj}, \texttt{k\_proj}, \texttt{v\_proj}, \texttt{o\_proj}) were targeted with a rank of 16 and a scaling factor of 16. The training is performed with a batch size of 2, gradient accumulation of 4 steps, and a learning rate of $1 \times 10^{-4}$ using the AdamW optimizer and capped at 25 epochs. Mixed precision (FP16/BF16) accelerates training, while regular evaluations and early stopping ensure convergence. This optimized approach efficiently trains the models for generating canonical phrases and helping positives.

\begin{figure*}[h]
\centering
\small

\begin{subfigure}{0.49\textwidth}
  \centering
  \setlength{\lineskip}{2pt}
\includegraphics[width=0.32\linewidth]{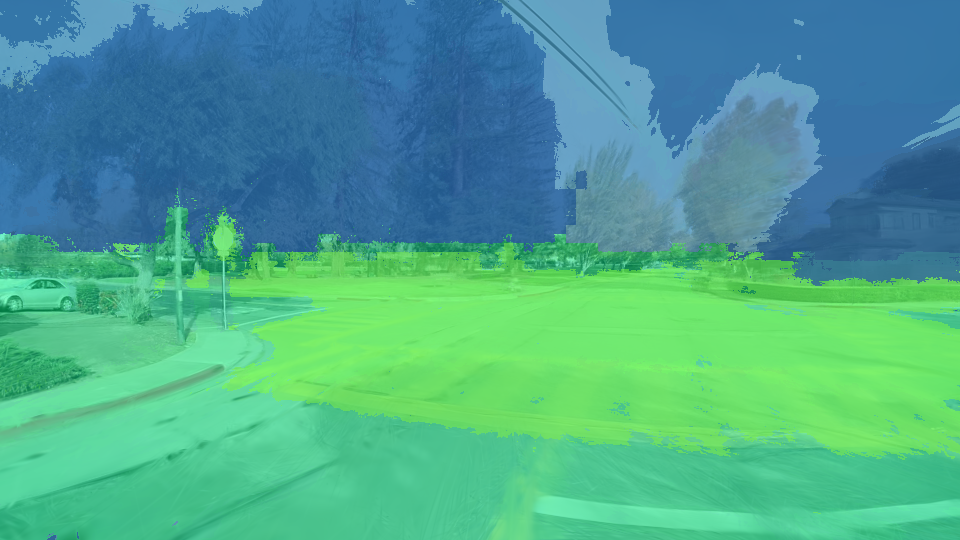}\hspace{0.002\linewidth}
\includegraphics[width=0.32\linewidth]{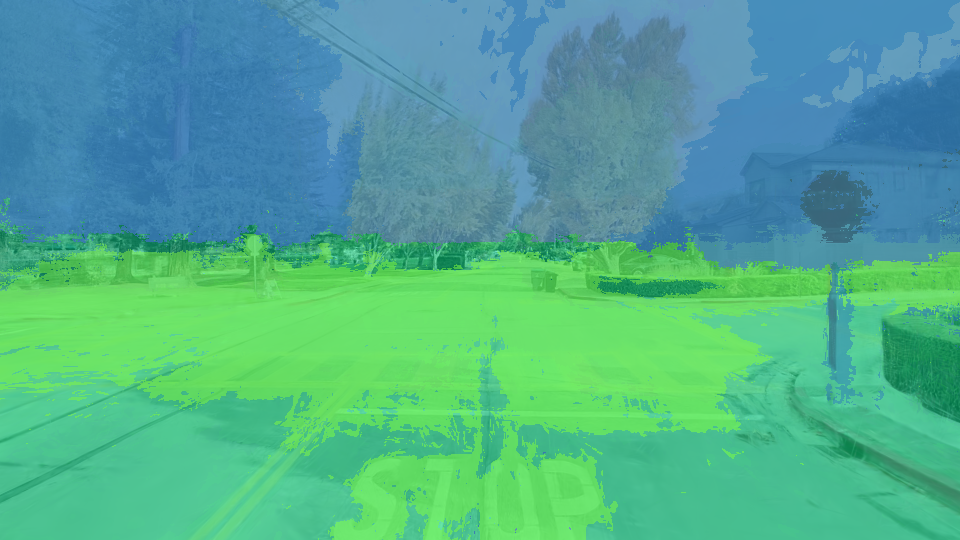}\hspace{0.002\linewidth}
\includegraphics[width=0.32\linewidth]{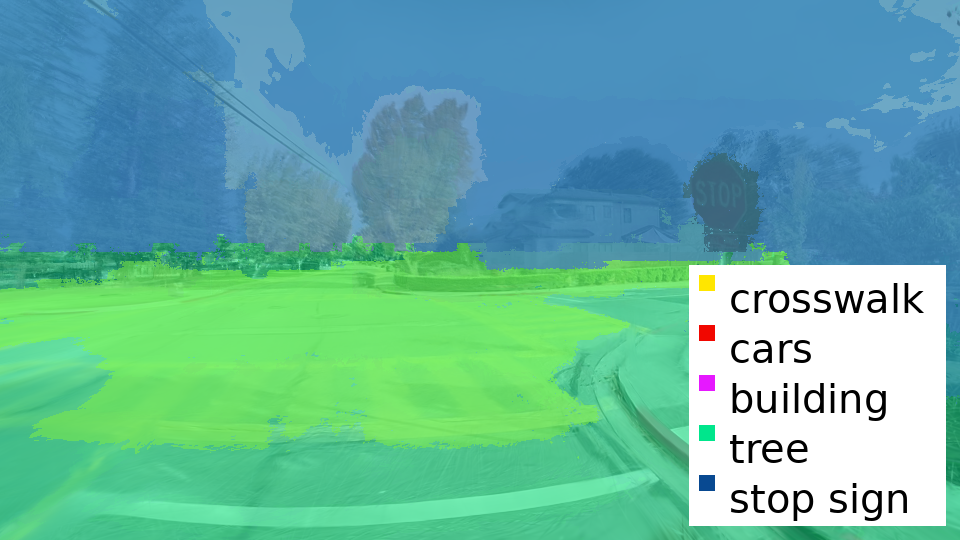}
\\
\includegraphics[width=0.32\linewidth]{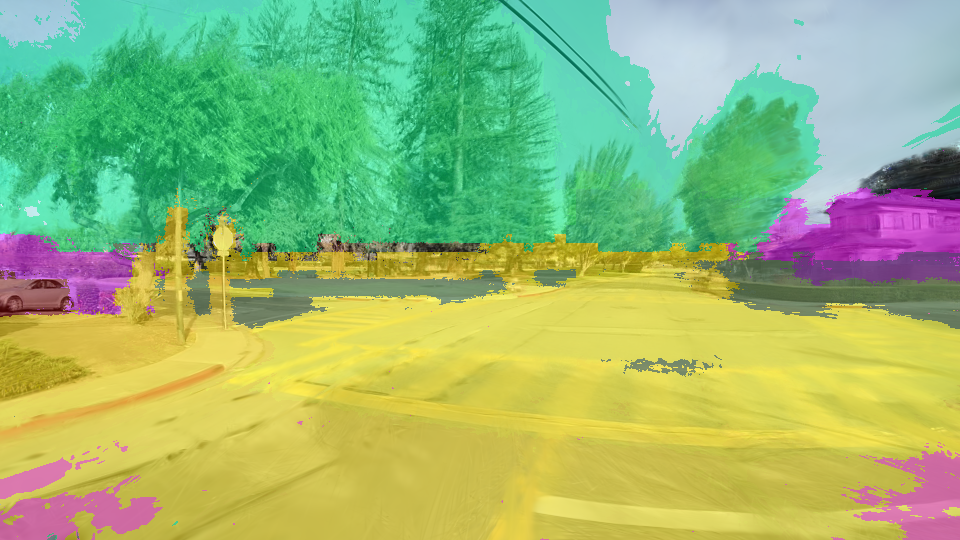}\hspace{0.002\linewidth}
\includegraphics[width=0.32\linewidth]{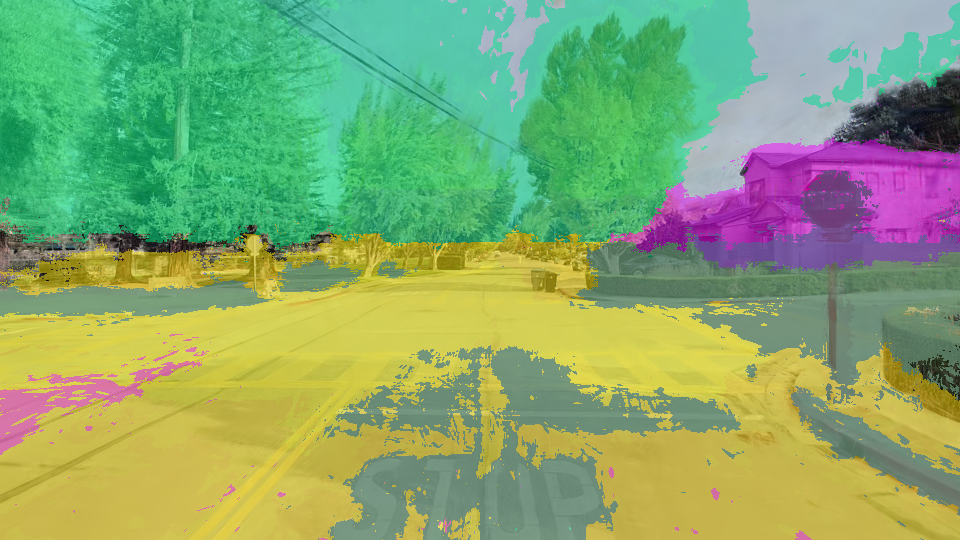}\hspace{0.002\linewidth}
\includegraphics[width=0.32\linewidth]{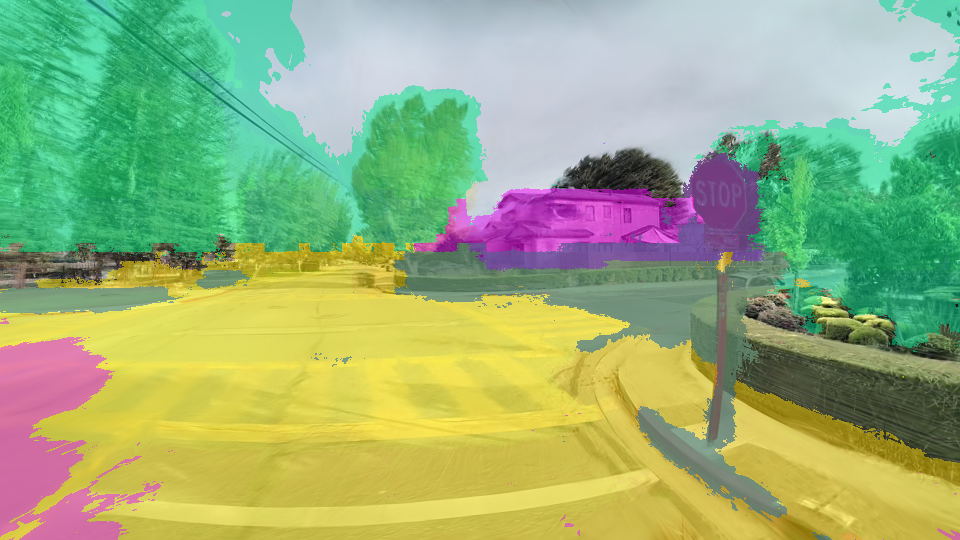}
\caption{Qwen 0.5B Instruct (top) and fine-tuned (bottom)}
\end{subfigure}
\hfill
\begin{subfigure}{0.49\textwidth}
  \centering
  \setlength{\lineskip}{2pt}
\includegraphics[width=0.32\linewidth]{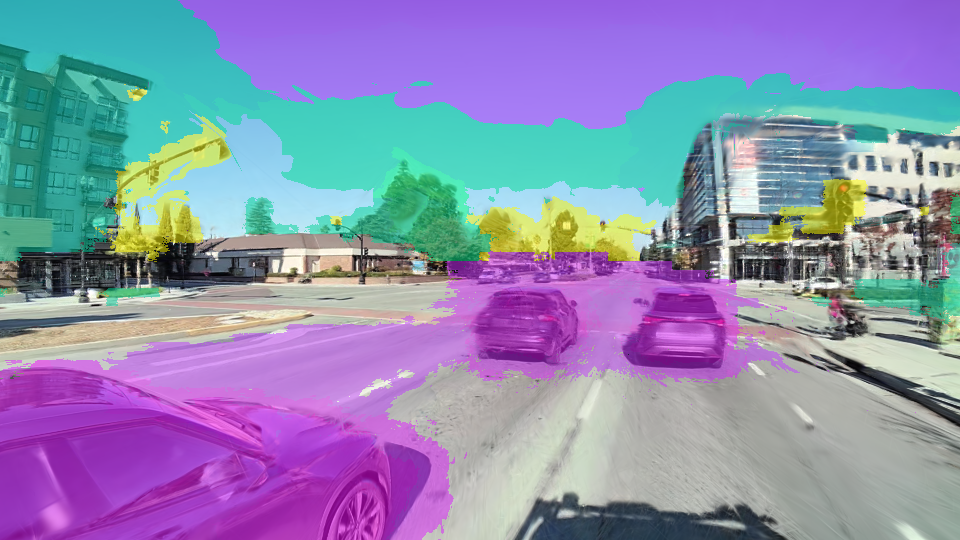}\hspace{0.002\linewidth}
\includegraphics[width=0.32\linewidth]{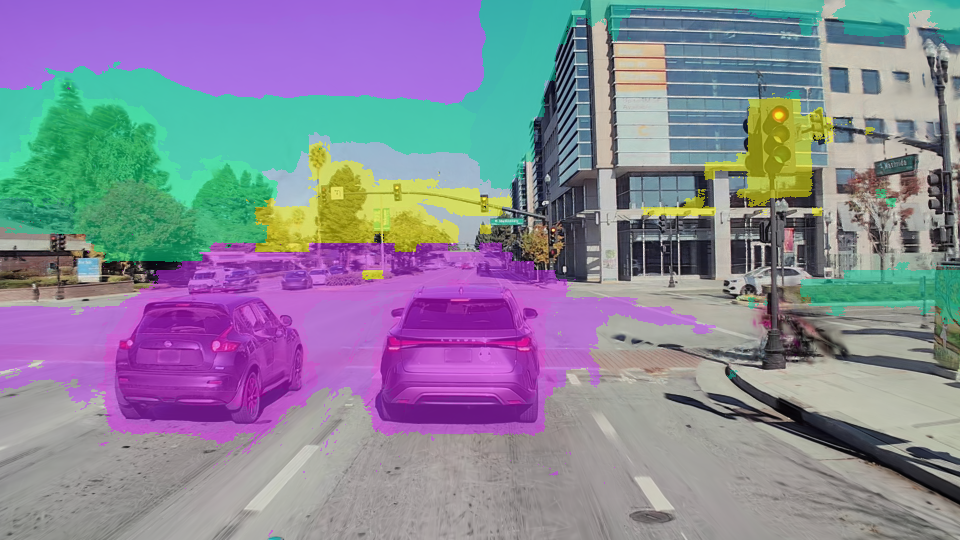}\hspace{0.002\linewidth}
\includegraphics[width=0.32\linewidth]{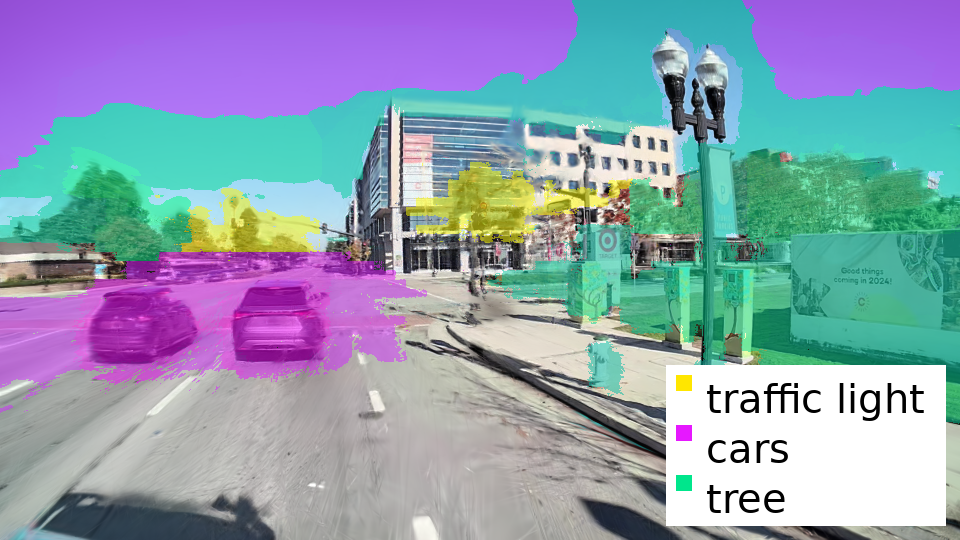}
\\
\includegraphics[width=0.32\linewidth]{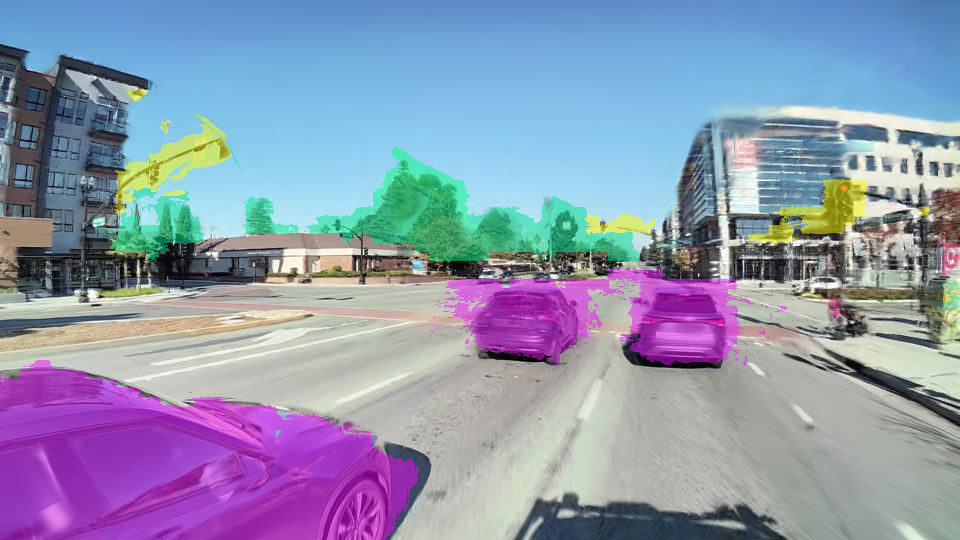}\hspace{0.002\linewidth}
\includegraphics[width=0.32\linewidth]{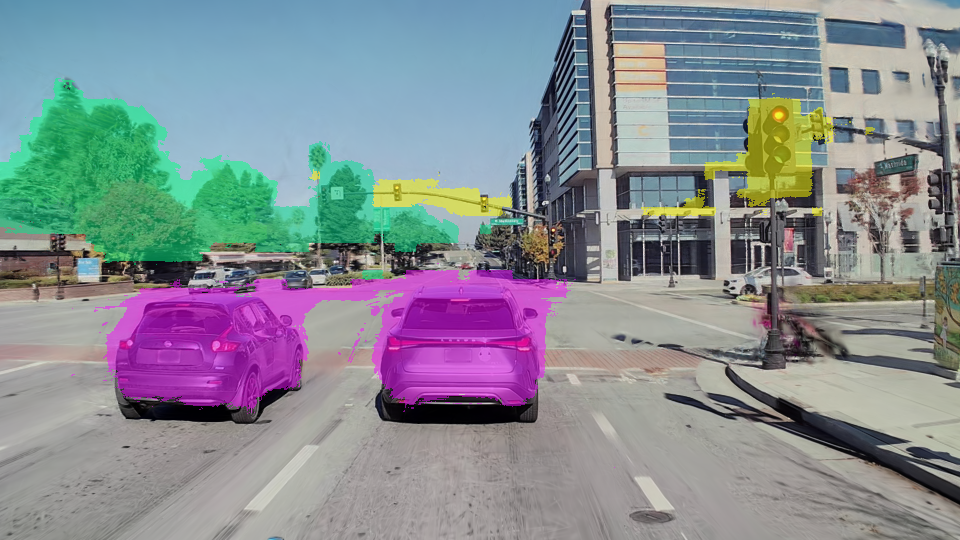}\hspace{0.002\linewidth}
\includegraphics[width=0.32\linewidth]{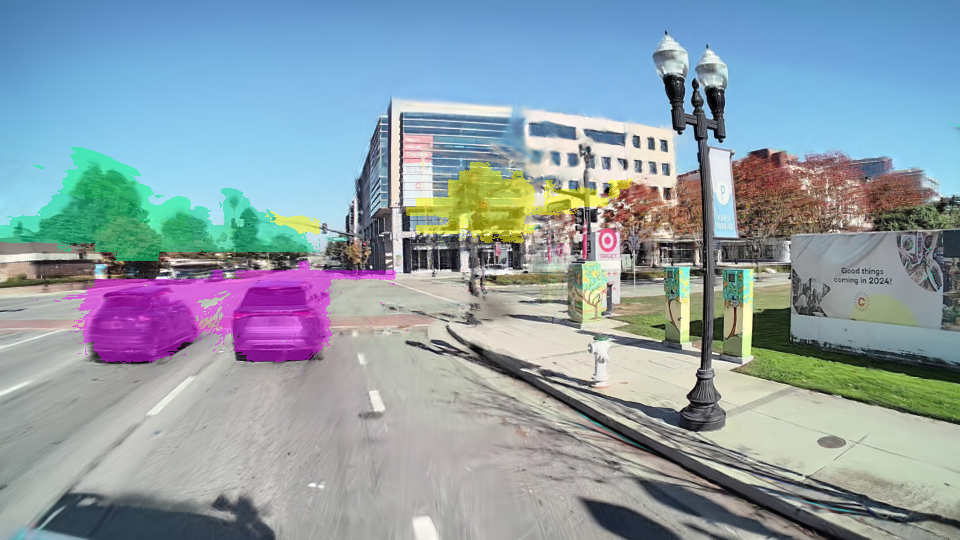}
\caption{Qwen 1.5B Instruct (top) and fine-tuned (bottom)}
\end{subfigure}

\begin{subfigure}{0.49\textwidth}
  \centering
  \setlength{\lineskip}{2pt}
\includegraphics[width=0.32\linewidth]{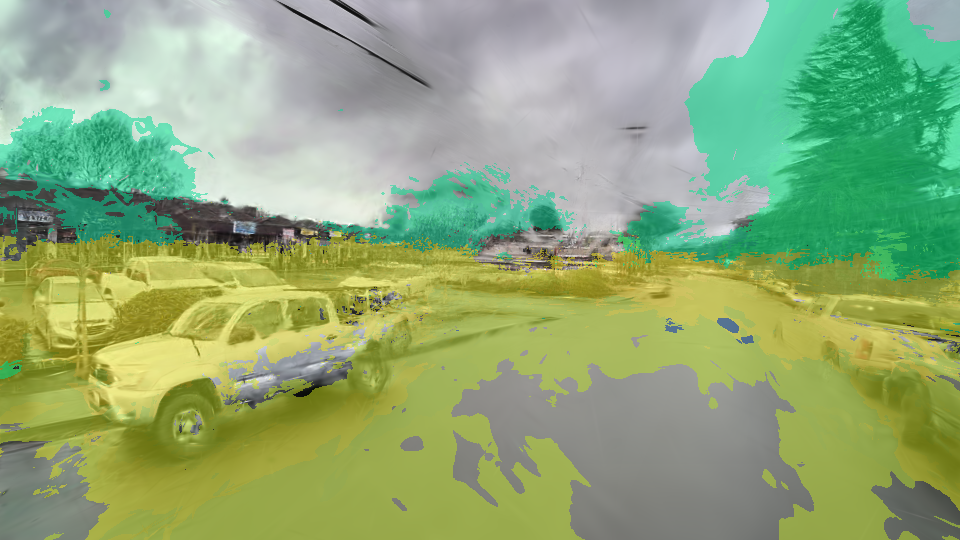}\hspace{0.002\linewidth}
\includegraphics[width=0.32\linewidth]{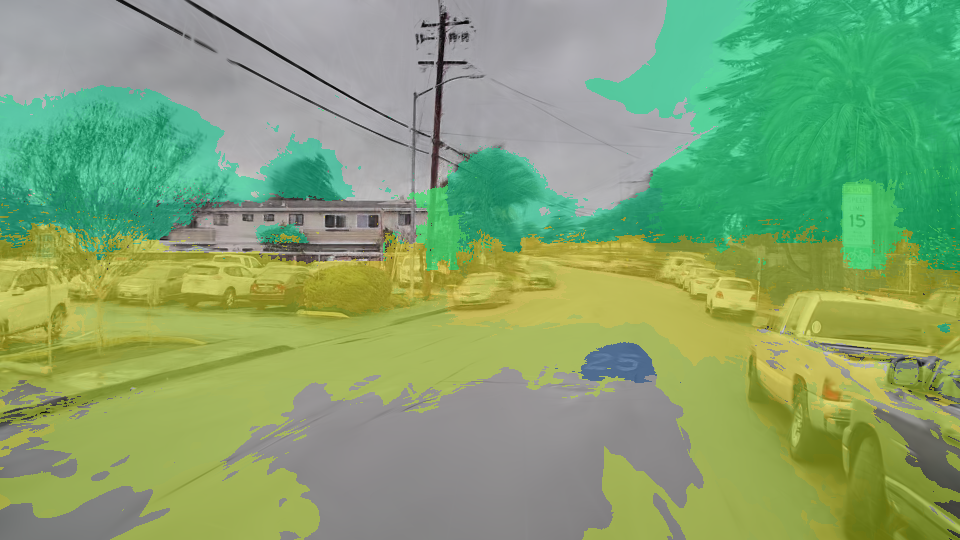}\hspace{0.002\linewidth}
\includegraphics[width=0.32\linewidth]{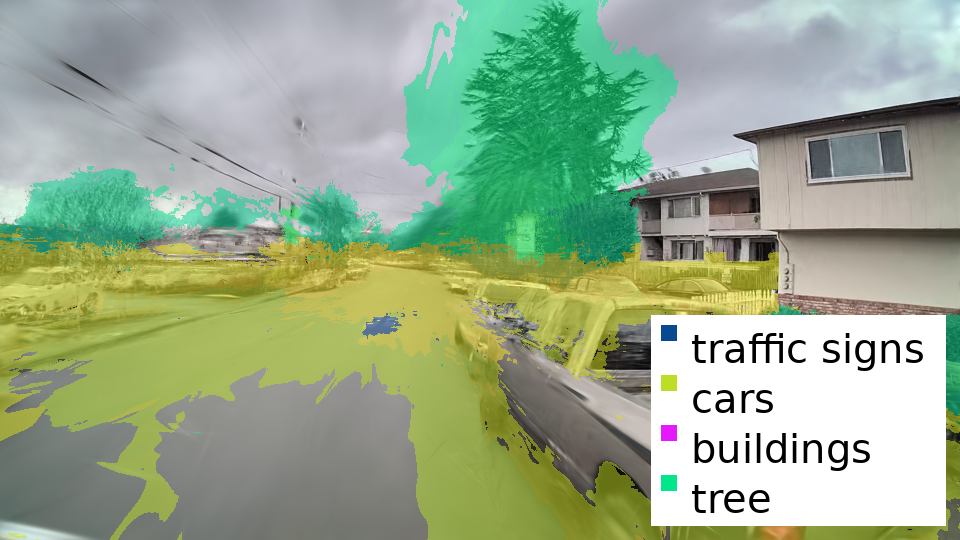}
\\
\includegraphics[width=0.32\linewidth]{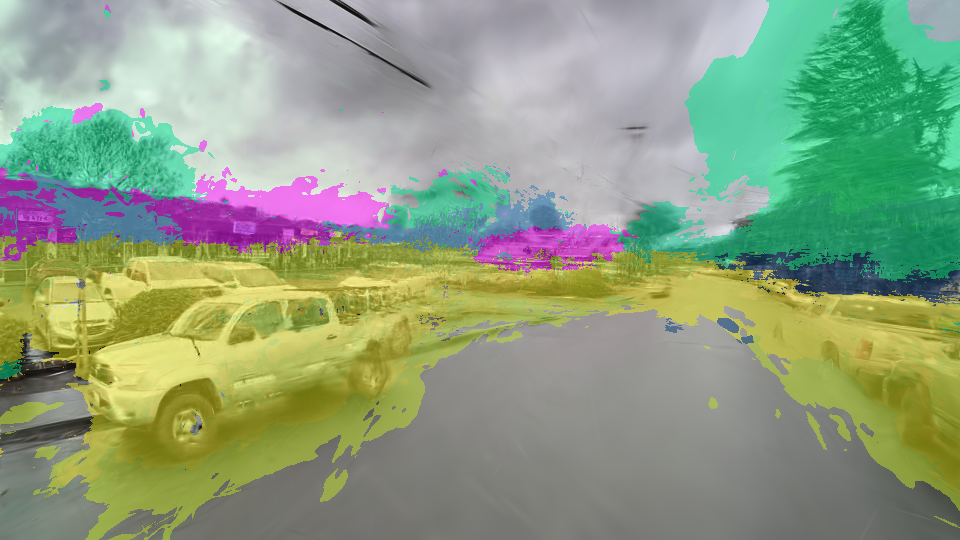}\hspace{0.002\linewidth}
\includegraphics[width=0.32\linewidth]{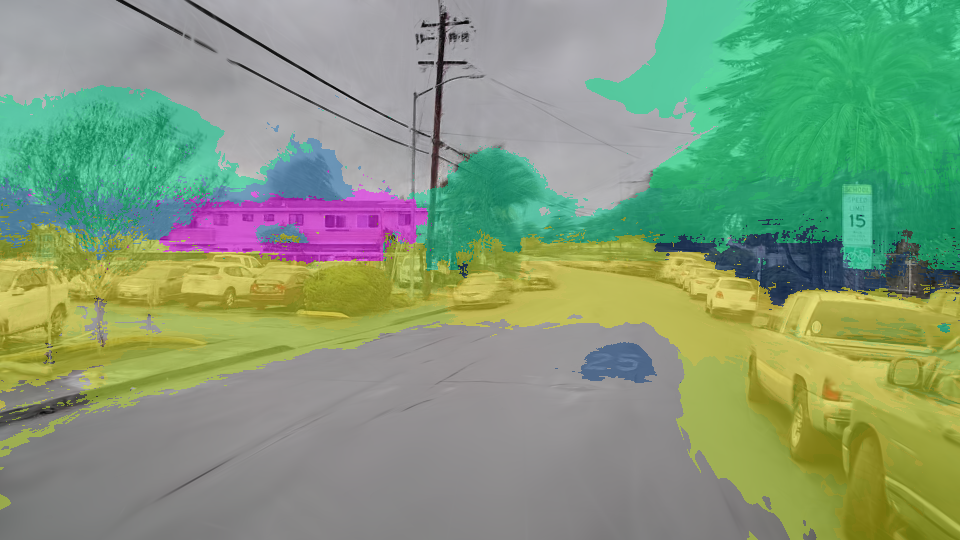}\hspace{0.002\linewidth}
\includegraphics[width=0.32\linewidth]{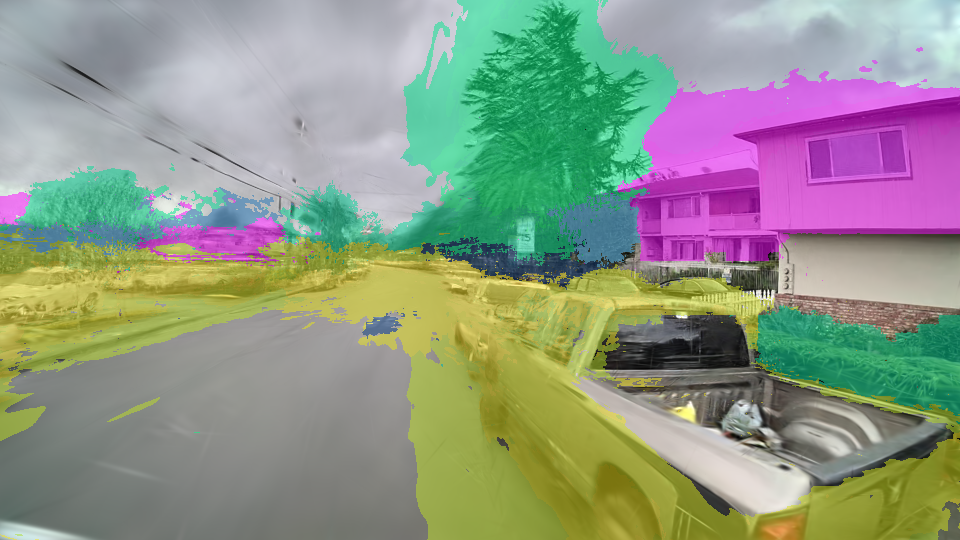}
\caption{Qwen 3B Instruct (top) and fine-tuned (bottom)}
\end{subfigure}
\hfill
\begin{subfigure}{0.49\textwidth}
  \centering
  \setlength{\lineskip}{2pt}
\includegraphics[width=0.32\linewidth]{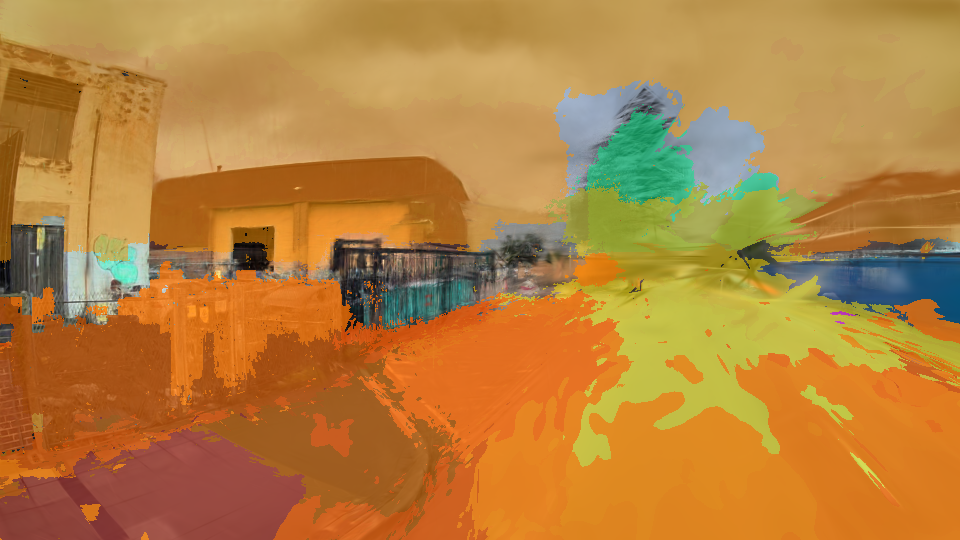}\hspace{0.002\linewidth}
\includegraphics[width=0.32\linewidth]{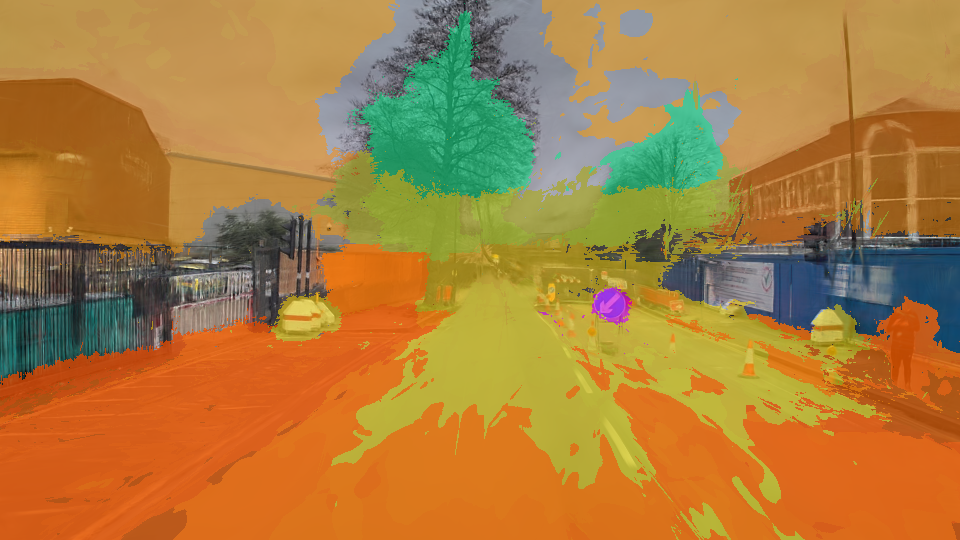}\hspace{0.002\linewidth}
\includegraphics[width=0.32\linewidth]{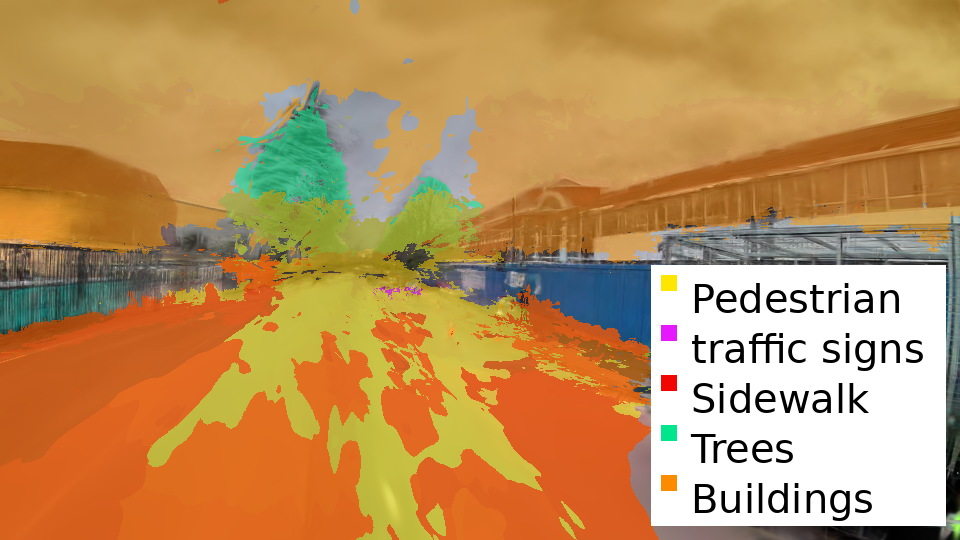}
\\
\includegraphics[width=0.32\linewidth]{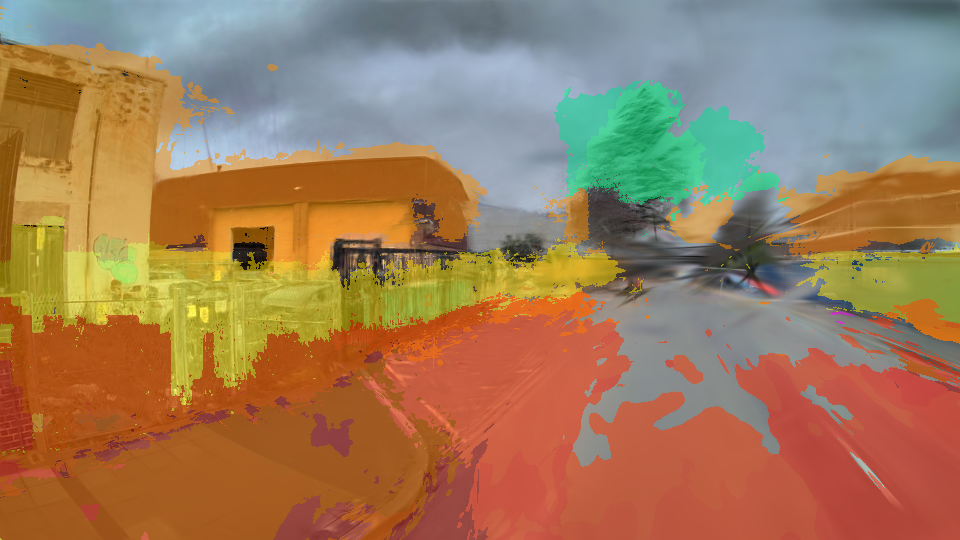}\hspace{0.002\linewidth}
\includegraphics[width=0.32\linewidth]{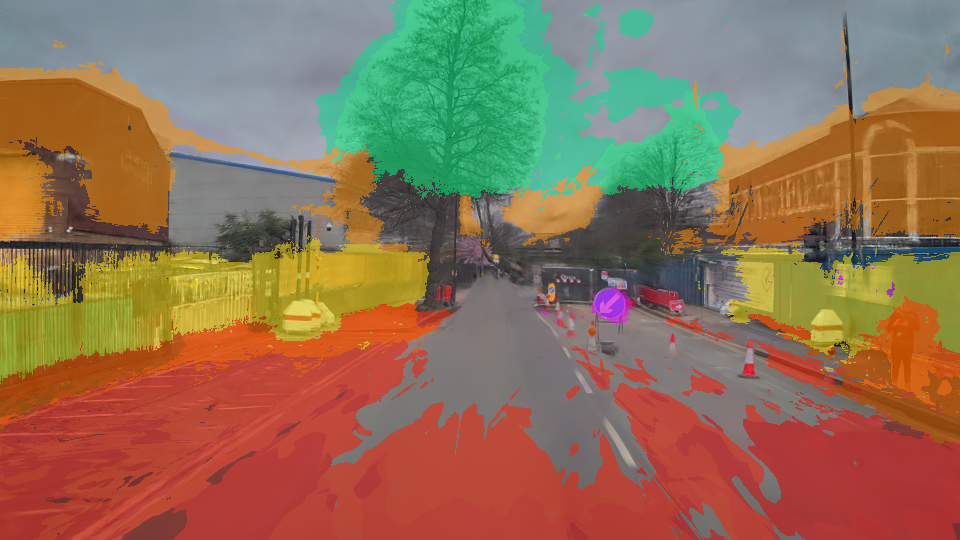}\hspace{0.002\linewidth}
\includegraphics[width=0.32\linewidth]{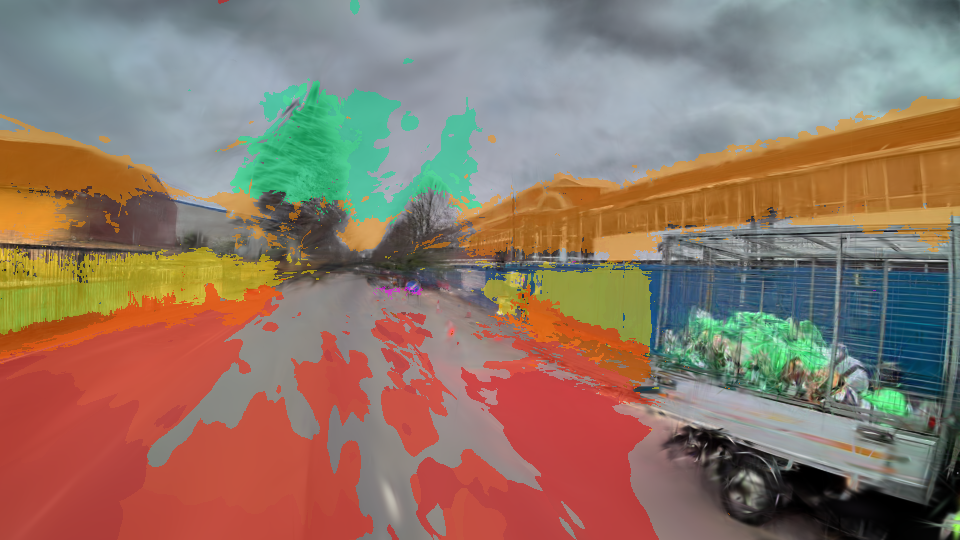}

\caption{Qwen 7B Instruct (top), fine-tuned (bottom)}
\end{subfigure}
\begin{subfigure}{0.49\textwidth}
  \centering
  \setlength{\lineskip}{2pt}
\includegraphics[width=0.32\linewidth]{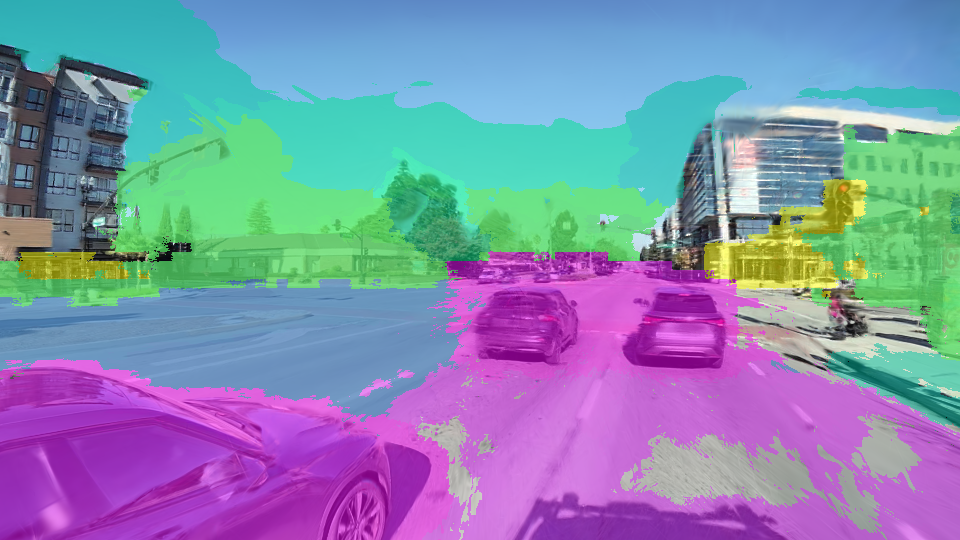}\hspace{0.002\linewidth}
\includegraphics[width=0.32\linewidth]{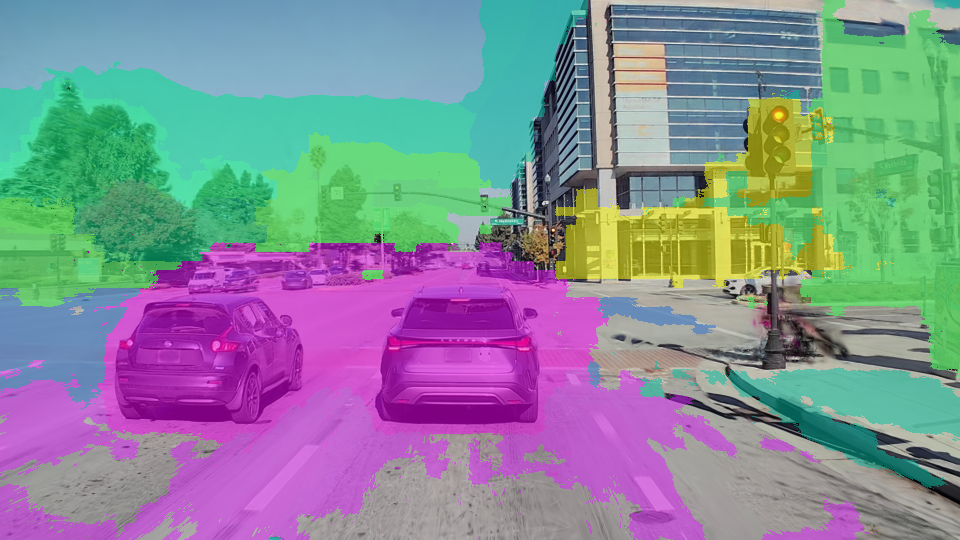}\hspace{0.002\linewidth}
\includegraphics[width=0.32\linewidth]{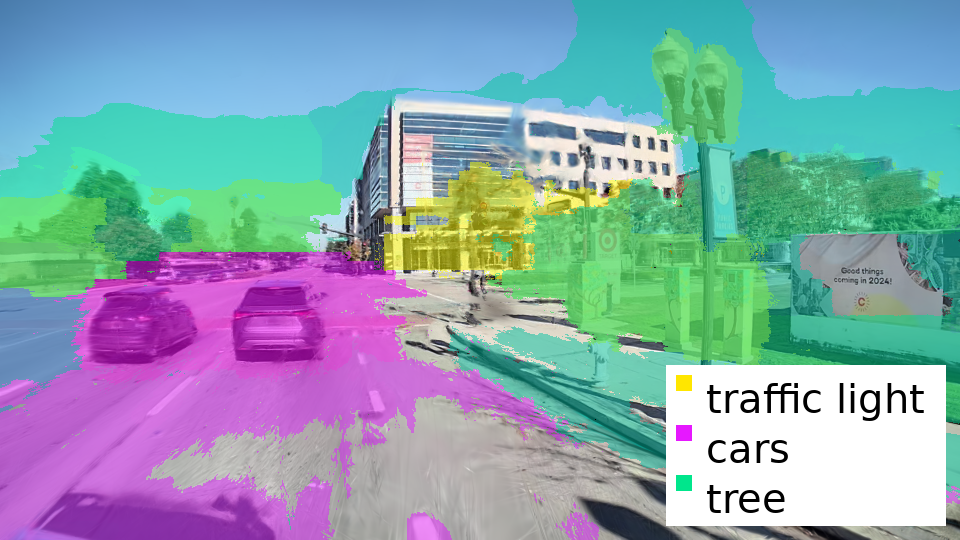}
\\
\includegraphics[width=0.32\linewidth]{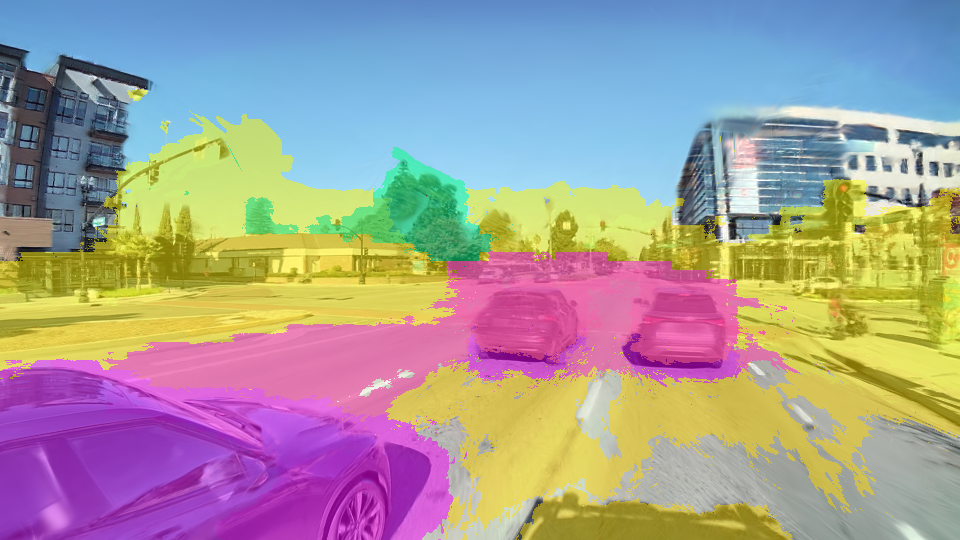}\hspace{0.002\linewidth}
\includegraphics[width=0.32\linewidth]{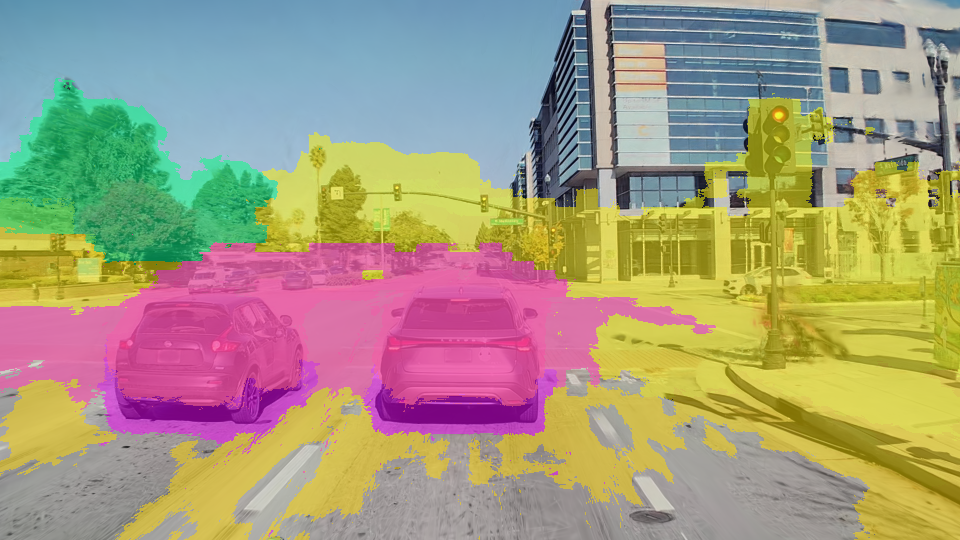}\hspace{0.002\linewidth}
\includegraphics[width=0.32\linewidth]{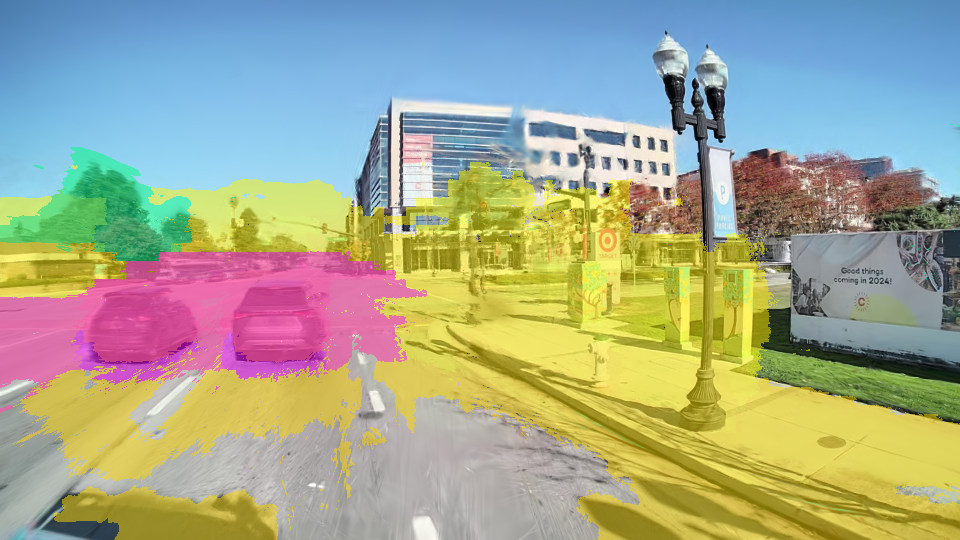}
\caption{Llama 1B (top) and fine-tuned (bottom)}
\end{subfigure}
\hfill
\begin{subfigure}{0.49\textwidth}
  \centering
  \setlength{\lineskip}{2pt}
\includegraphics[width=0.32\linewidth]{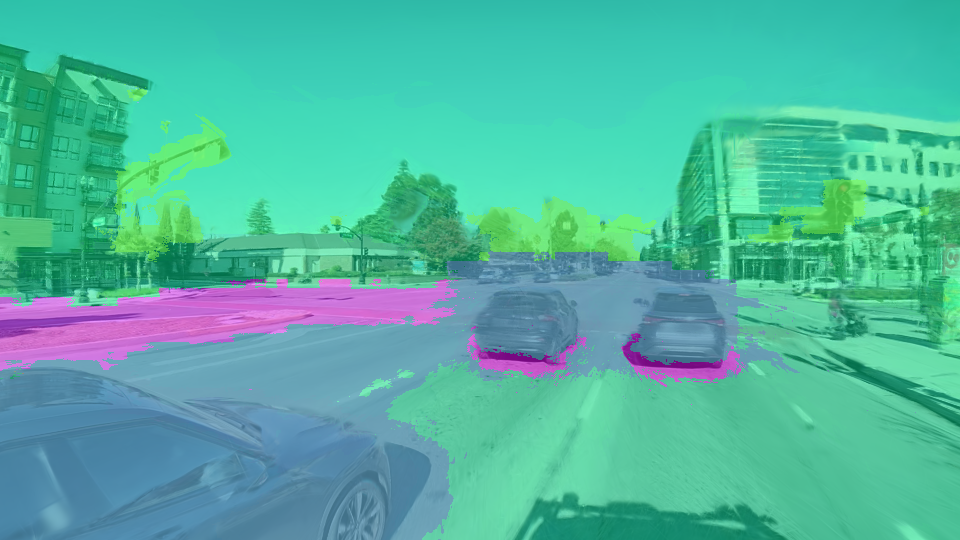}\hspace{0.002\linewidth}
\includegraphics[width=0.32\linewidth]{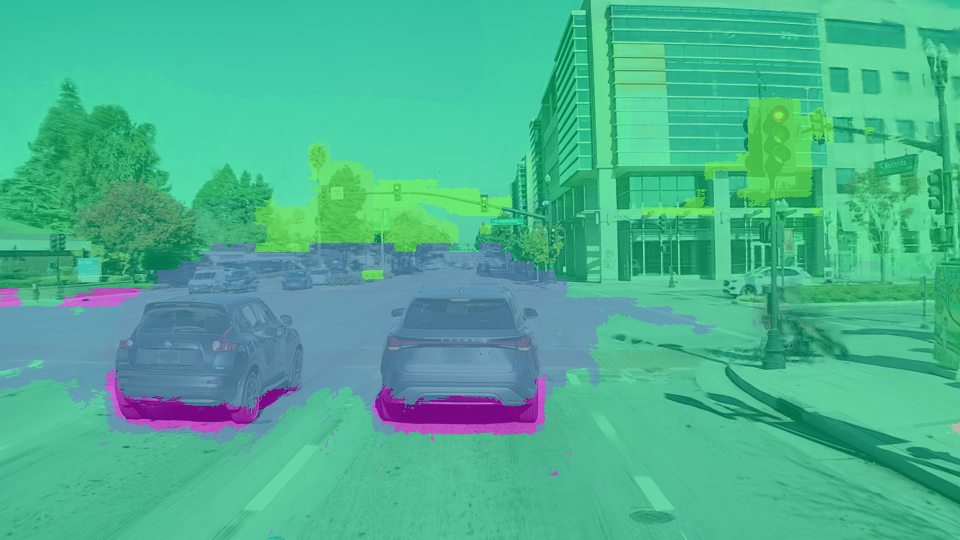}\hspace{0.002\linewidth}
\includegraphics[width=0.32\linewidth]{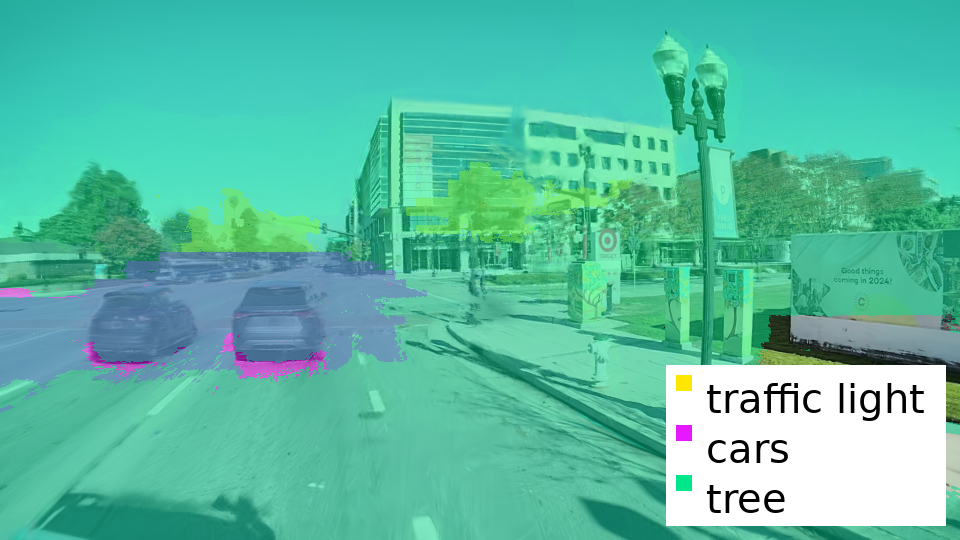}
\\
\includegraphics[width=0.32\linewidth]{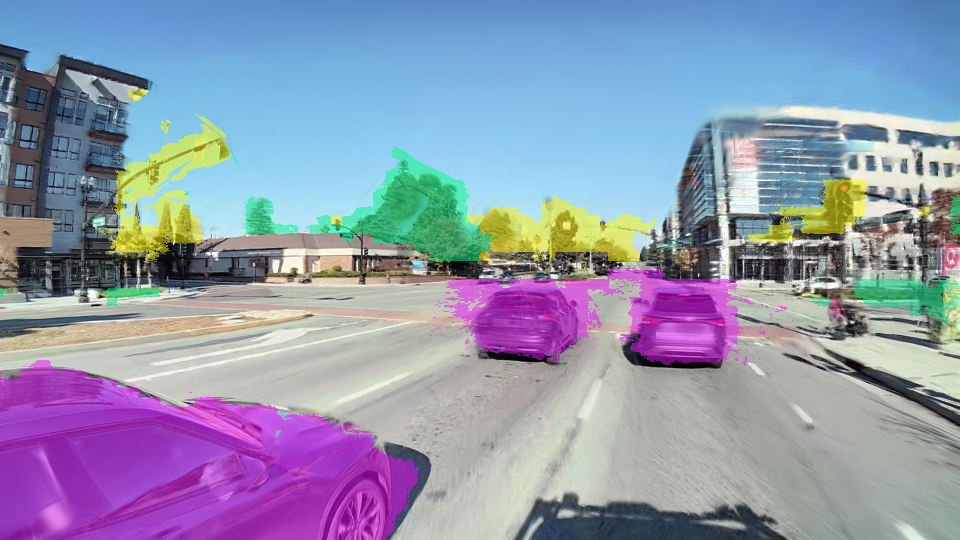}\hspace{0.002\linewidth}
\includegraphics[width=0.32\linewidth]{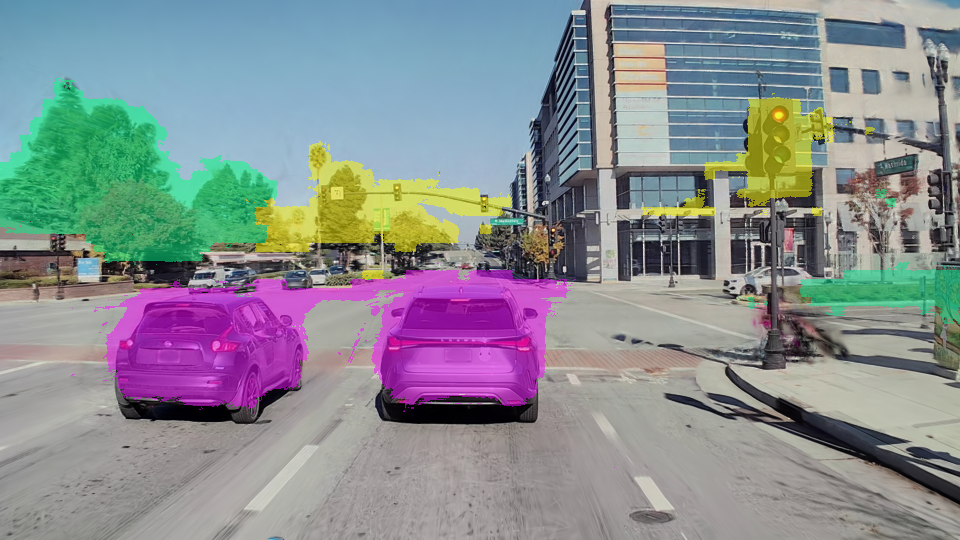}\hspace{0.002\linewidth}
\includegraphics[width=0.32\linewidth]{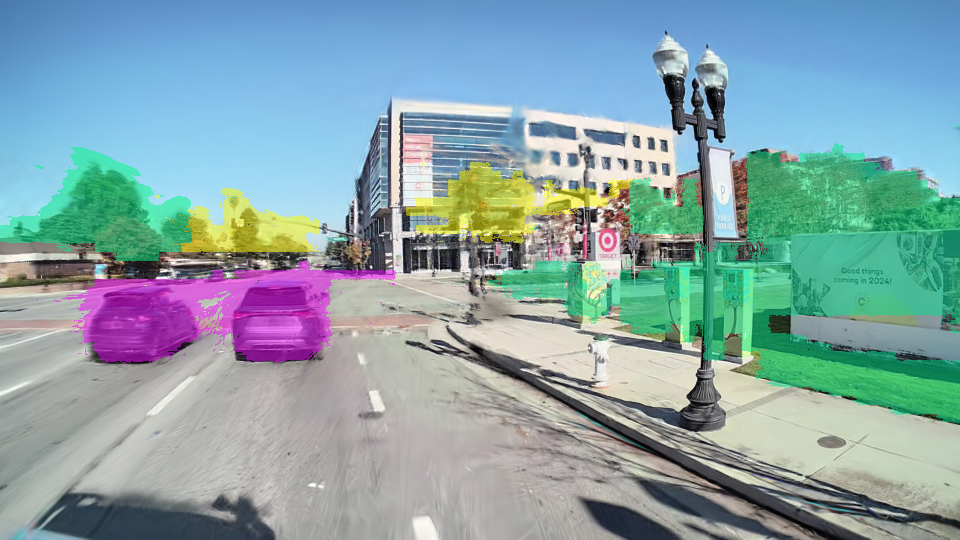}
\caption{Llama 3B (top) and fine-tuned (bottom)}
\end{subfigure}

\caption{Qualitative comparison of semantic segmentation results before and after fine-tuning for different model variants. Each subfigure shows three views of the same scene, with the top row representing results from the instruction-tuned model (before fine-tuning) and the bottom row showing results from the fine-tuned model.}
\label{fig:comparison}
\end{figure*}

\subsection{Implementation}
We train the LE3DGS using the same loss function as described in \cite{shi2023le3dgaussians} on the Wayvesene101 dataset. We set $\lambda_{s}=0.5 $, $\lambda_{CE}=\lambda_{smo}=\lambda_{u}=0.1$ and other weights to the default. The training process involved two main phases: dense semantic feature extraction and 3D Gaussian point optimization. For the semantic feature extraction, we set the number of epochs to 500, ensuring a rich semantic representation of the scene. In the optimization phase for 3D Gaussian points, we employed a 2D position gradient threshold of 0.0001, with a total of 20,000 iterations. The densification process was halted halfway through, at 10,000 iterations. All other hyperparameters were kept consistent with those used in \cite{kerbl20233dgaussiansplattingrealtime}. The entire training process was conducted on a single RTX 4090 GPU, taking approximately 40 minutes to complete for each scene in the dataset. The mean peak signal-to-noise ratio (PSNR) achieved is around 25.14.

For the inference phase, we leveraged the API of GPT-3.5 Turbo~\cite{openai2024gpt35} and the small Instruct-tuned and fine-tuned Qwen and Llama models running on the device to generate queries, helping positives and canonical words dynamically.
We then calculated the relevancy score using Algorithm \ref{alg:relevancy}. We applied a threshold of 0.5 to the computed relevancy scores to obtain object segmentations. The canonical phrases used in the baseline approach are “object”, “things”, “stuff”, and “texture”, consistent with the approach described in~\cite{shi2023le3dgaussians}.

\subsection{Results}
We evaluated the performance of Qwen and Llama model variants and GPT-3.5 Turbo against the baseline approach for open vocabulary segmentation through Intersection over Union (IoU), Accuracy, Precision, and mean Average Precision (mAP) metrics.
Our experimental results, presented in Table~\ref{tab:performance}, demonstrate that GPT-3.5 Turbo significantly outperforms the baseline approach across all metrics, achieving an IoU of $0.27$, accuracy of $0.90$, and precision of $0.37$ without helping positives, compared to the baseline's IoU of $0.18$, accuracy of $0.64$, and precision of $0.19$. This substantial improvement validates the effectiveness of leveraging large language models for open vocabulary segmentation tasks.

The fine-tuned variants of both Qwen and Llama models show marked improvements over their instruction-tuned counterparts. Notably, the Qwen2.5-1.5B-Finetuned model achieves the highest IoU ($0.28$) and precision ($0.37$) among all models in the configuration without helping positives, matching GPT-3.5 Turbo's performance. Similarly, the Llama3.2-3B-Finetuned model demonstrates strong performance with an IoU of $0.26$ and accuracy of $0.84$, suggesting that effective fine-tuning can elevate the performance of smaller models to the levels approaching that of much larger models. Figure~\ref{fig:comparison} shows the qualitative comparison of fine-tuned and instruct-tuned models.

In our ablation study examining the quality of generated helping positives ($P_{help-pos}$), we observe that a model's capacity significantly affects its ability to generate effective helping words. Larger models generate more semantically meaningful helping positives, leading to stable or improved segmentation performance, while smaller variants generate less effective helping words that can degrade segmentation quality. For instance, as shown in Table~\ref{tab:performance}, the helping positives generated by Qwen2.5-7B-Finetuned enable robust segmentation performance (IoU: $0.23$, Accuracy: $0.89$, Precision: $0.34$), while helping positives from smaller models lead to more significant performance drops in the segmentation pipeline.

\begin{figure*}[h]
  \centering
  \begin{minipage}{0.97\textwidth}
    \centering
    \begin{minipage}{00.24\textwidth}
      \centering
      \includegraphics[width=\linewidth]{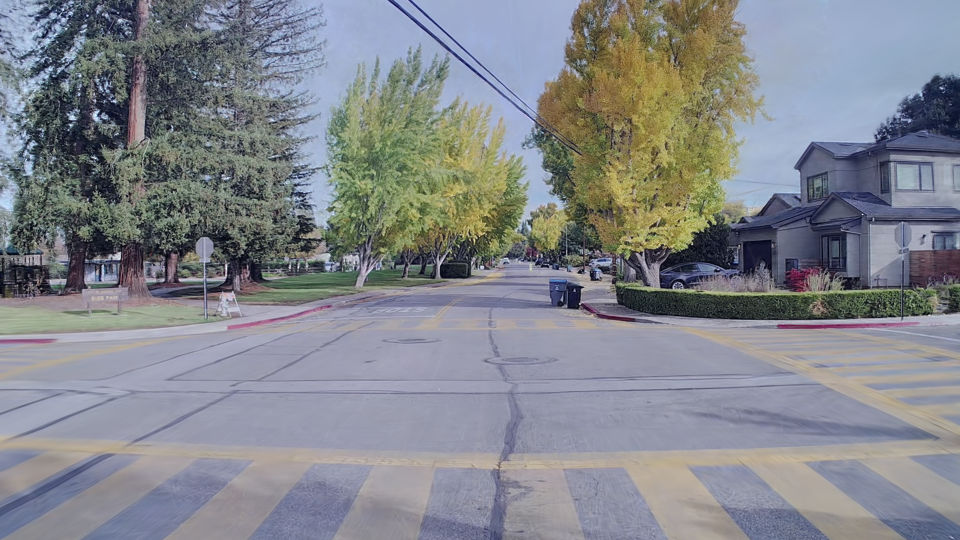}
      \par\smallskip 
    \end{minipage}
    \hfill
    \begin{minipage}{00.24\textwidth}
      \centering
      \includegraphics[width=\linewidth]{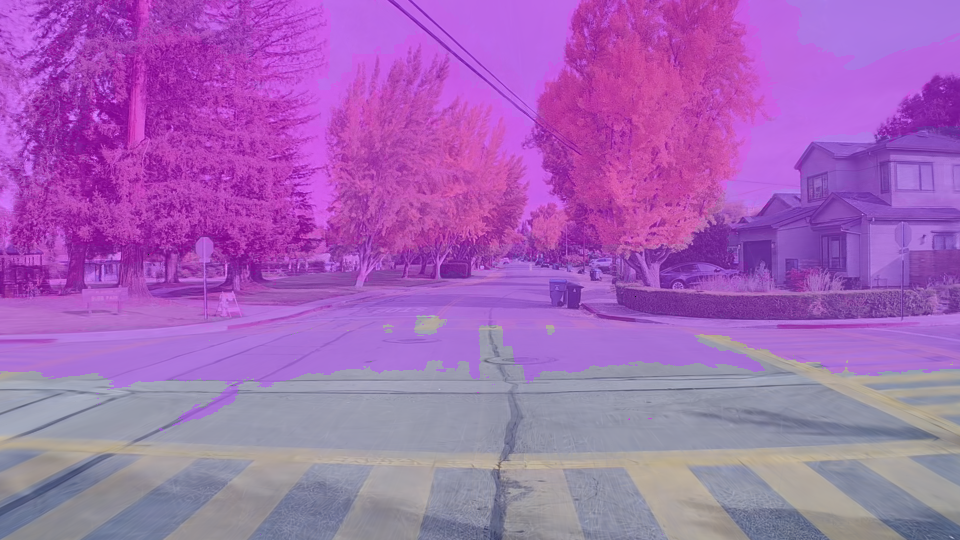}
      \par\smallskip 
    \end{minipage}
    \hfill
    \begin{minipage}{00.24\textwidth}
      \centering
      \includegraphics[width=\linewidth]{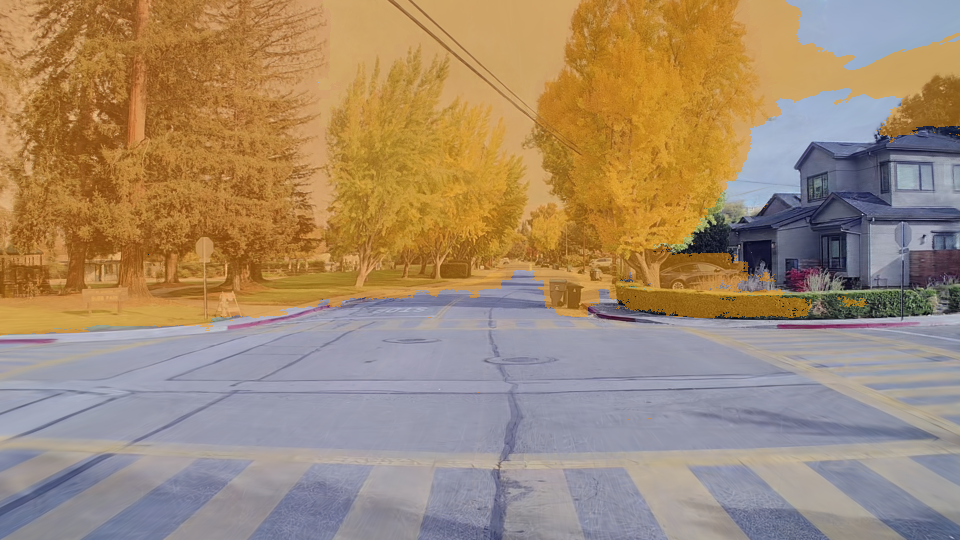}
      \par\smallskip 
    \end{minipage}
    \hfill
    \begin{minipage}{00.24\textwidth}
      \centering
      \includegraphics[width=\linewidth]{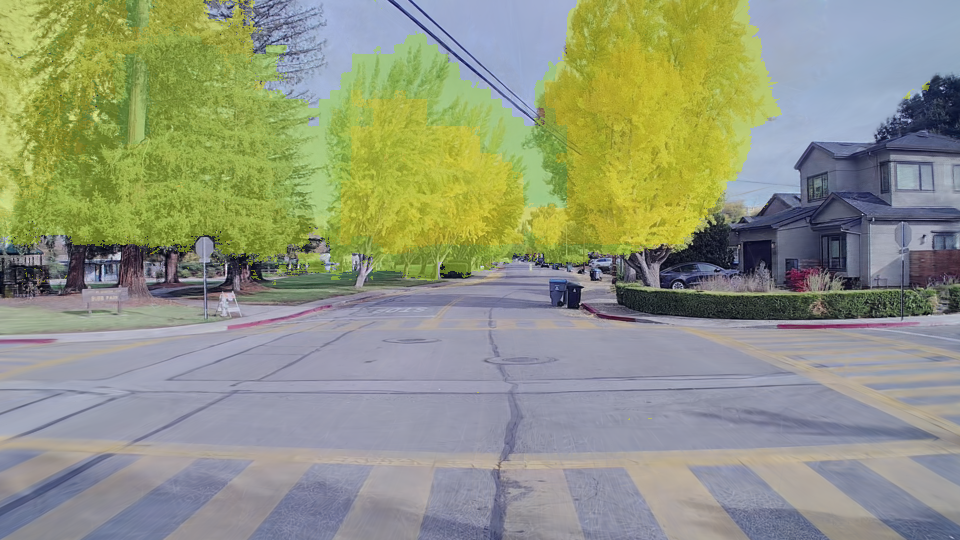}
      \par\smallskip 
    \end{minipage}
  \end{minipage}
  
  \begin{minipage}{0.97\textwidth}
    \centering
    \begin{minipage}{00.24\textwidth}
      \centering
      \includegraphics[width=\linewidth]{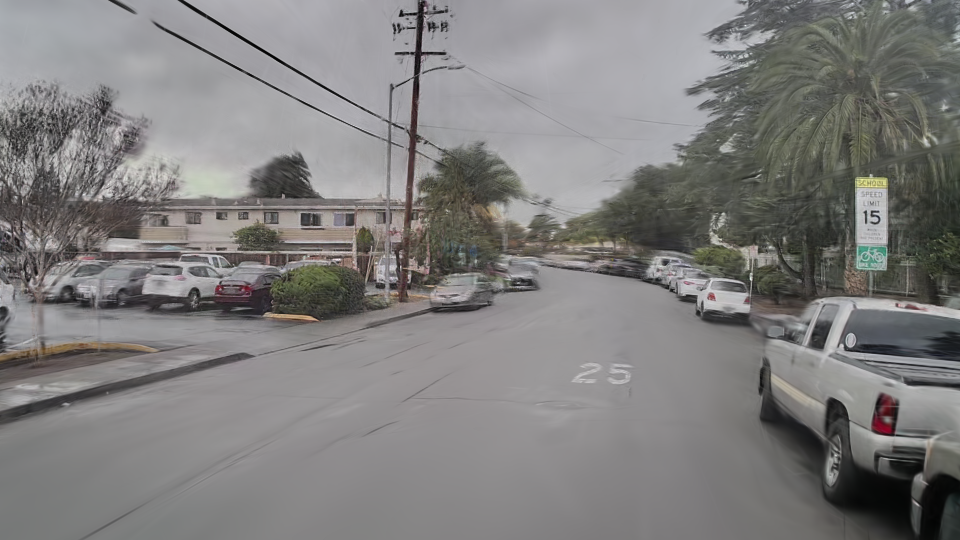}
      \par\smallskip 
    \end{minipage}
    \hfill
    \begin{minipage}{00.24\textwidth}
      \centering
      \includegraphics[width=\linewidth]{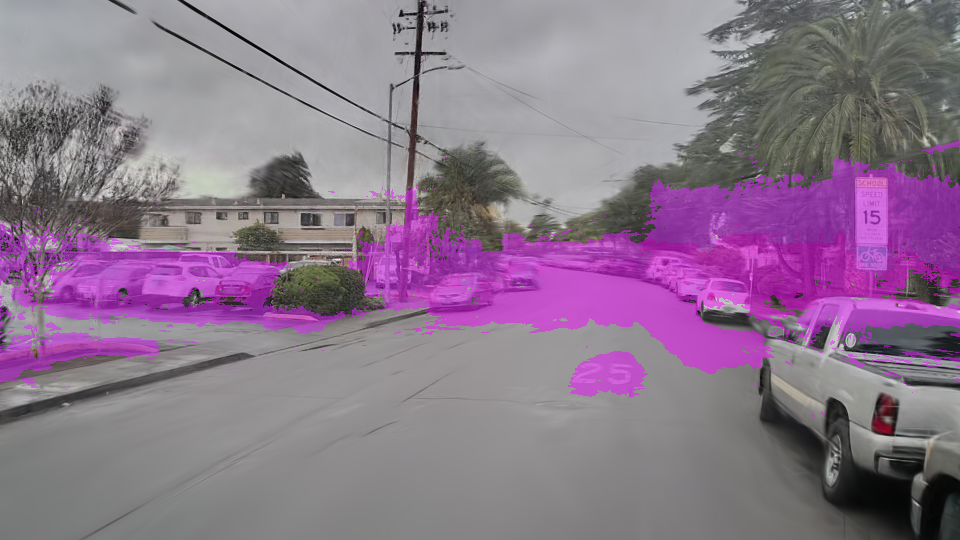}
      \par\smallskip 
    \end{minipage}
    \hfill
    \begin{minipage}{00.24\textwidth}
      \centering
      \includegraphics[width=\linewidth]{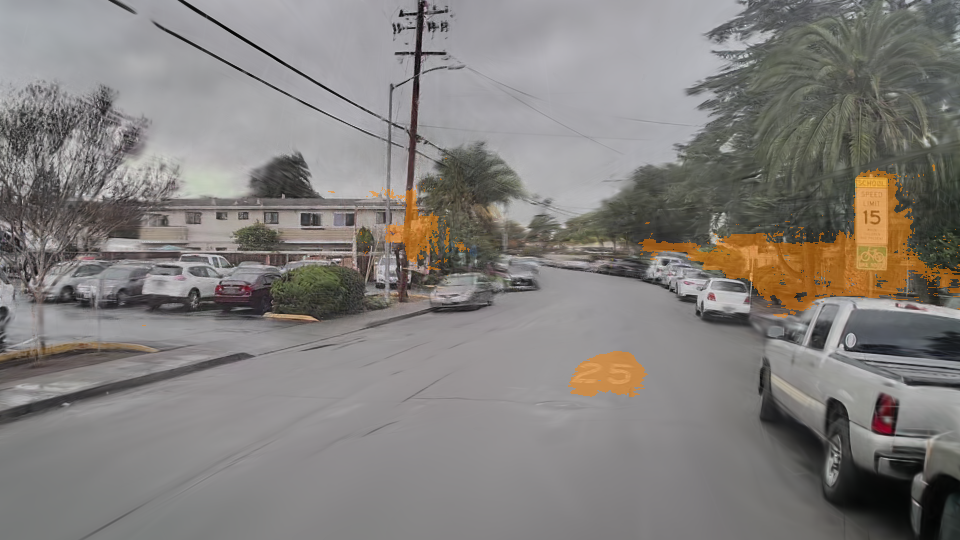}
      \par\smallskip 
    \end{minipage}
    \hfill
    \begin{minipage}{00.24\textwidth}
      \centering
      \includegraphics[width=\linewidth]{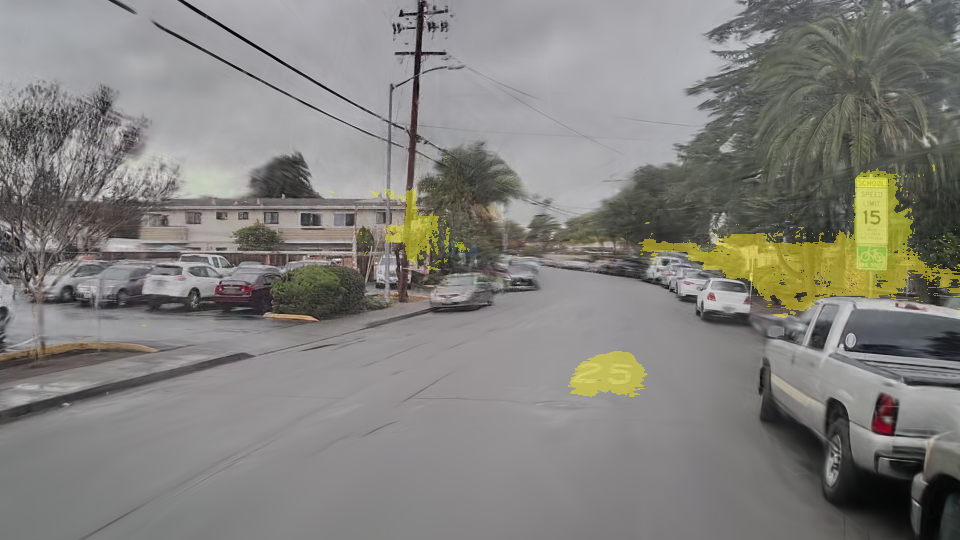}
      \par\smallskip 
    \end{minipage}
  \end{minipage}
  
  \begin{minipage}{0.97\textwidth}
    \centering
    \begin{minipage}{00.24\textwidth}
      \centering
      \includegraphics[width=\linewidth]{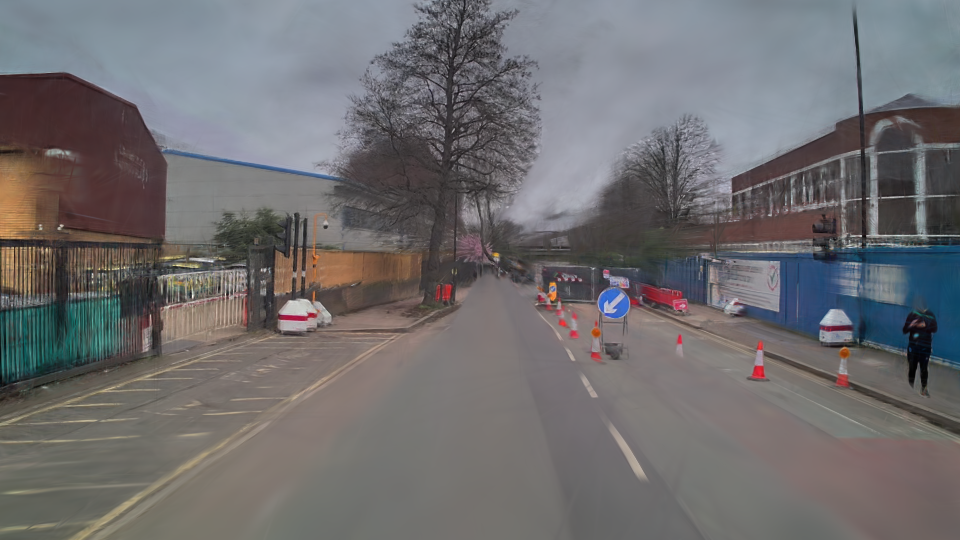}
      \par\smallskip 
    \end{minipage}
    \hfill
    \begin{minipage}{00.24\textwidth}
      \centering
      \includegraphics[width=\linewidth]{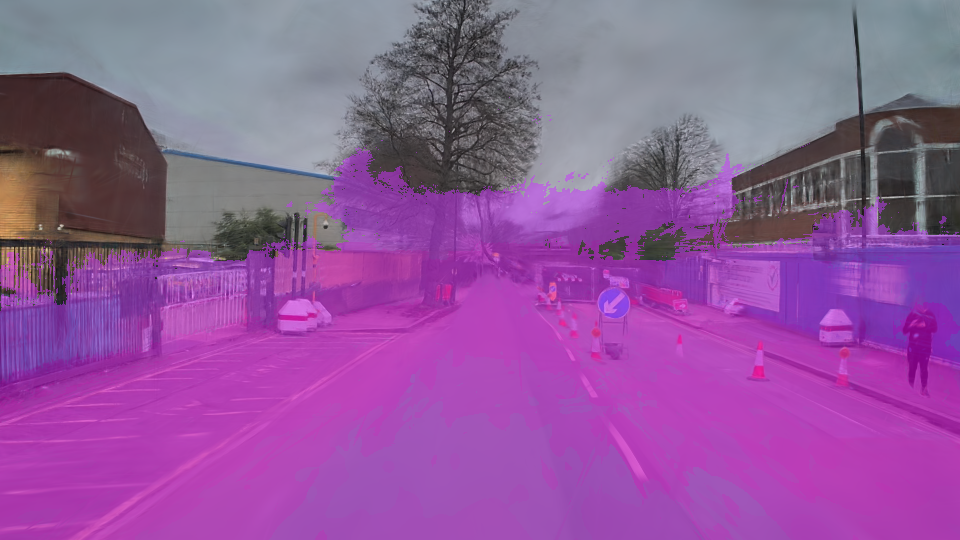}
      \par\smallskip 
    \end{minipage}
    \hfill
    \begin{minipage}{00.24\textwidth}
      \centering
      \includegraphics[width=\linewidth]{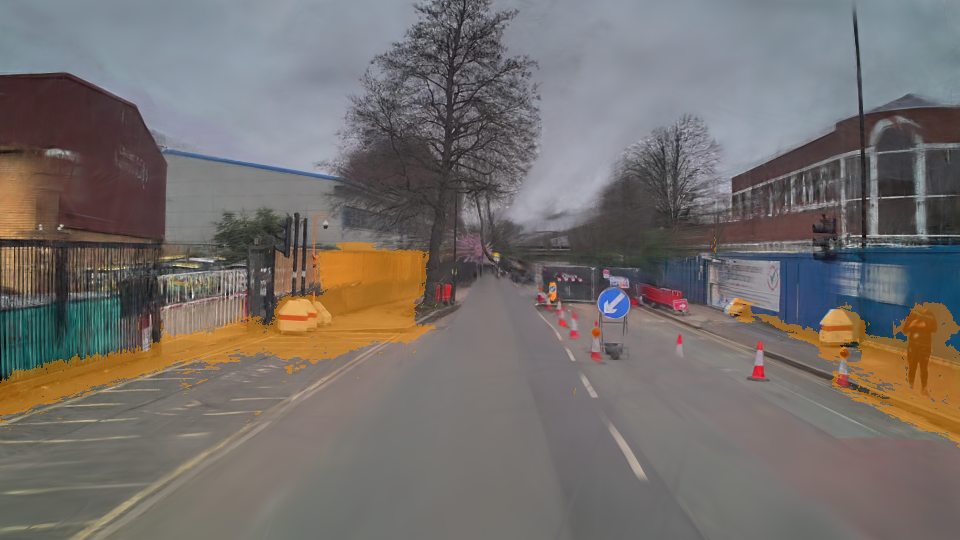}
      \par\smallskip 
    \end{minipage}
    \hfill
    \begin{minipage}{00.24\textwidth}
      \centering
      \includegraphics[width=\linewidth]{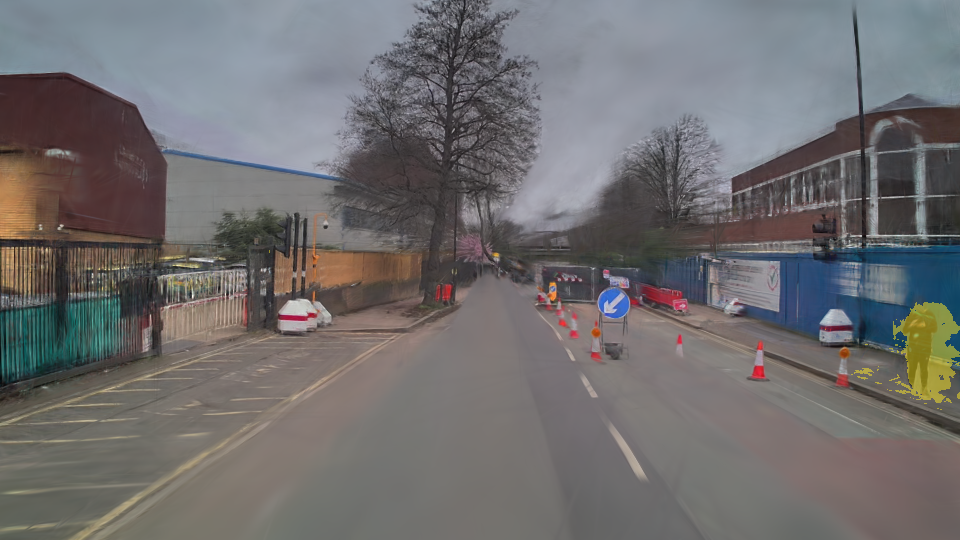}
      \par\smallskip 
    \end{minipage}
  \end{minipage}
  
  \begin{minipage}{0.97\textwidth}
    \centering
    \begin{minipage}{00.24\textwidth}
      \centering
      \includegraphics[width=\linewidth]{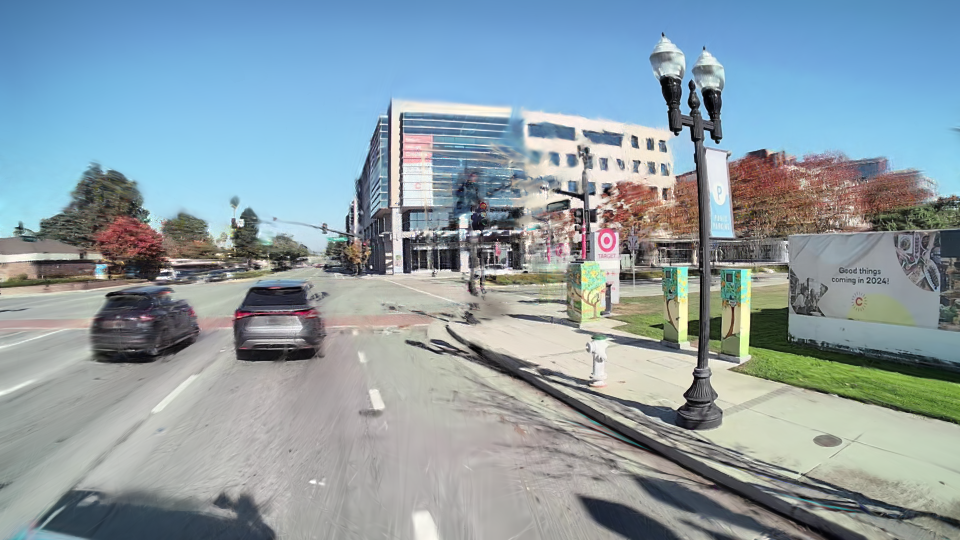}
      \par\smallskip 
    \end{minipage}
    \hfill
    \begin{minipage}{00.24\textwidth}
      \centering
      \includegraphics[width=\linewidth]{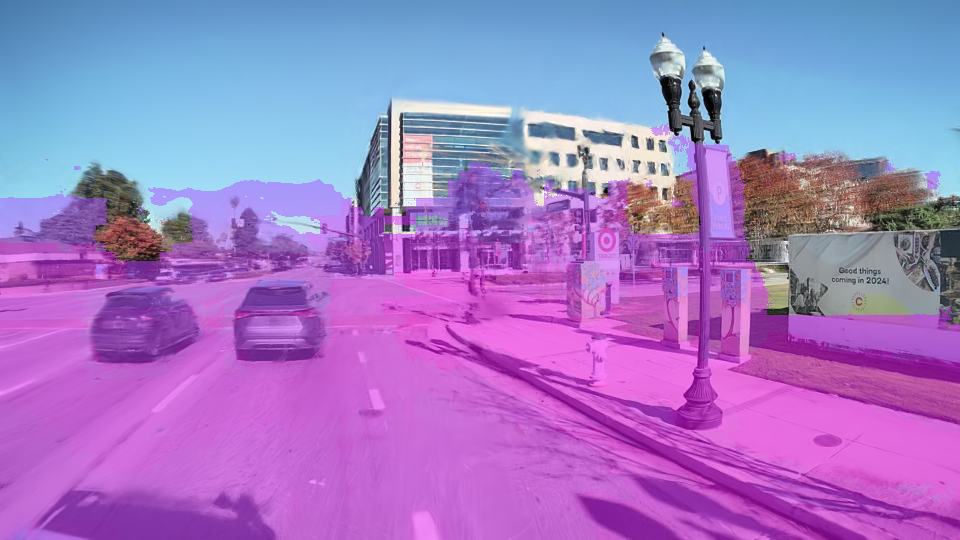}
      \par\smallskip 
    \end{minipage}
    \hfill
    \begin{minipage}{00.24\textwidth}
      \centering
      \includegraphics[width=\linewidth]{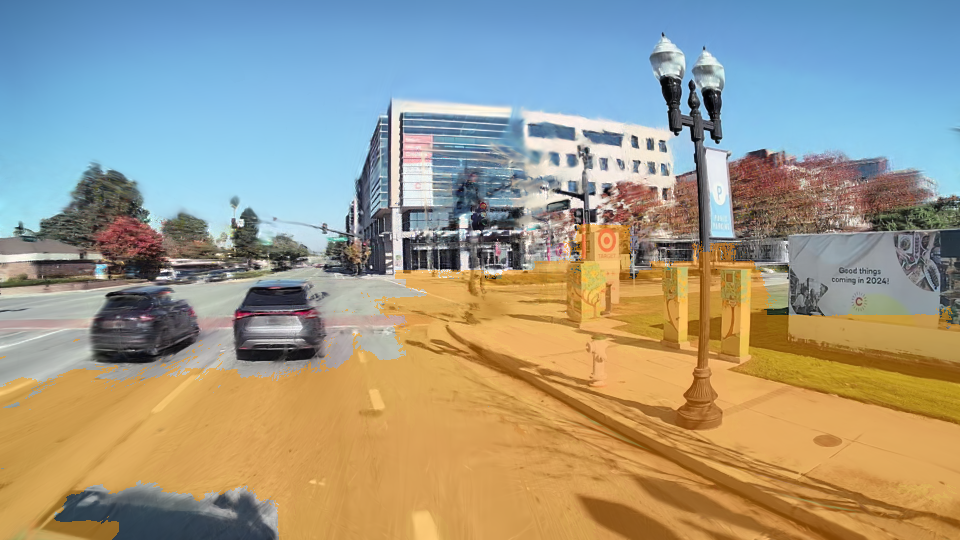}
      \par\smallskip 
    \end{minipage}
    \hfill
    \begin{minipage}{00.24\textwidth}
      \centering
      \includegraphics[width=\linewidth]{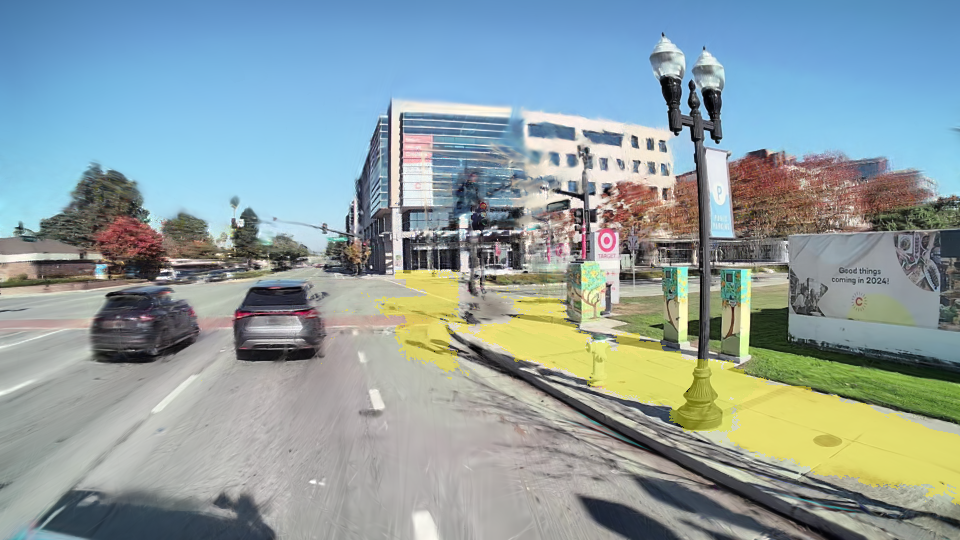}
      \par\smallskip 
    \end{minipage}
  \end{minipage}
  
  \begin{minipage}{0.97\textwidth}
    \centering
    \begin{minipage}{0.24\textwidth}
      \centering
      \includegraphics[width=\linewidth]{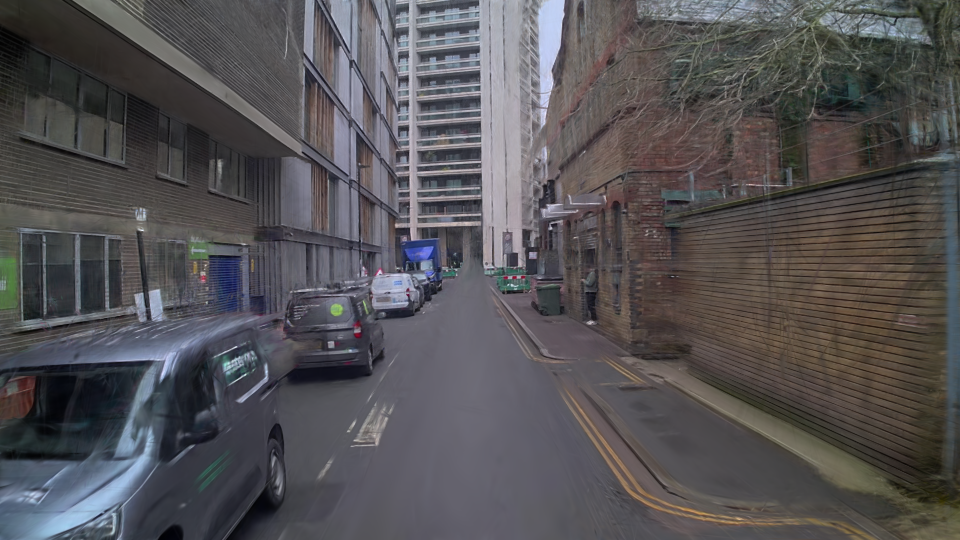}
      \par\smallskip 
      Reconstructed 
    \end{minipage}
    \hfill
    \begin{minipage}{0.24\textwidth}
      \centering
      \includegraphics[width=\linewidth]{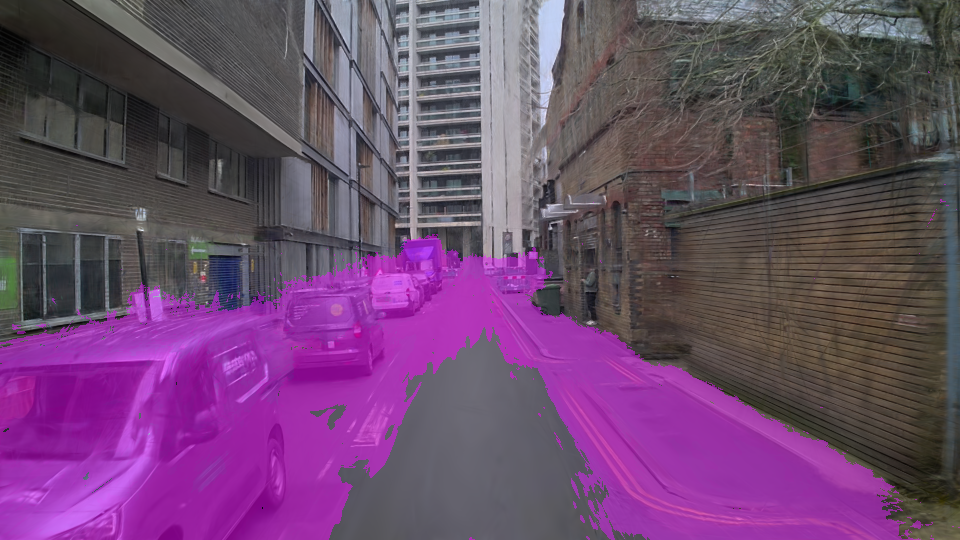}
      \par\smallskip 
      Predefined $p_{\texttt{canon}}$ 
    \end{minipage}
    \hfill
    \begin{minipage}{0.24\textwidth}
      \centering
      \includegraphics[width=\linewidth]{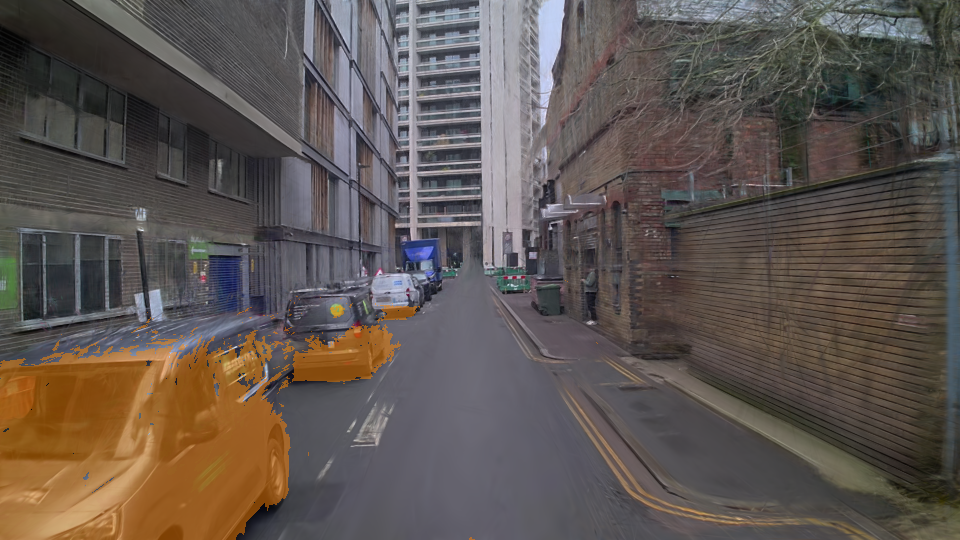}
      \par\smallskip 
      GPT without $p_{\texttt{help-pos}}$
    \end{minipage}
    \hfill
    \begin{minipage}{0.24\textwidth}
      \centering
      \includegraphics[width=\linewidth]{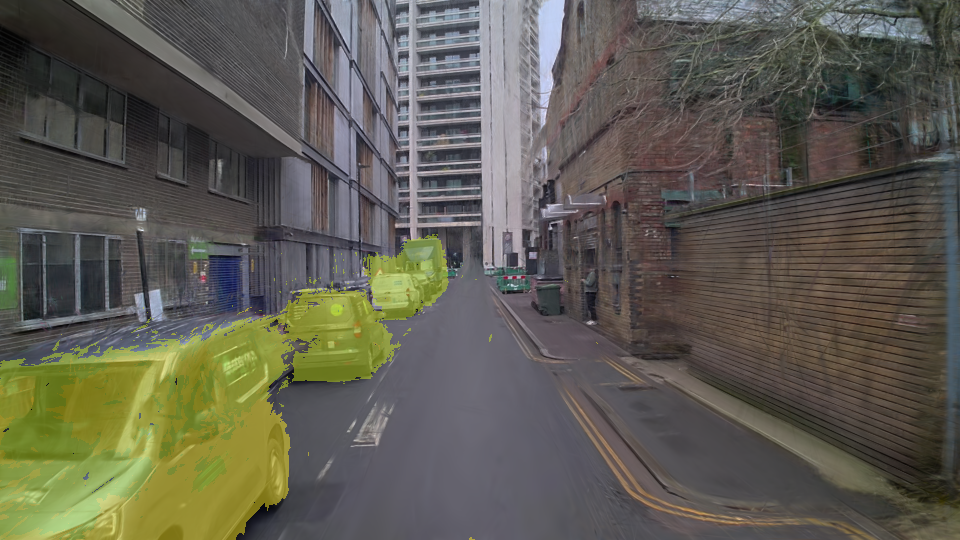}
      \par\smallskip 
      GPT with $p_{\texttt{help-pos}}$ 
    \end{minipage}
  \end{minipage}

  \caption{Visual comparison of query relevance across different scenes with different approaches. From top to bottom query words are: “trees”, “traffic signs”, “pedestrian”, “sidewalk”, and “cars”.}
  \label{fig:viscomp}
\end{figure*}

The quality of helping positives reaches its peak with GPT-3.5 Turbo, whose generated helping words enable substantial improvements in segmentation performance, achieving IoU ($0.32$, $+18.5\%$), accuracy ($0.94$, $+4.4\%$), and precision ($0.50$, $+35.1\%$). This marked enhancement suggests that larger-scale models can generate more contextually appropriate and semantically rich helping positives, leading to more effective scene segmentation. The quantitative results are further supported by qualitative comparisons shown in Figure~\ref{fig:viscomp}, which demonstrate improved segmentation quality across various urban scene elements.

\section{Conclusion}
In this study, we proposed a novel method that integrates LLMs with LE3DGS to enhance scene understanding for autonomous vehicles. Our approach significantly improves open vocabulary object segmentation in complex driving environments, substantially outperforming traditional predefined canonical phrase-based approaches. Our results demonstrate that carefully fine-tuned smaller language models can achieve performance comparable to larger expert models while enabling on-device applications. Furthermore, our ablation studies revealed that the effectiveness of helping positives correlates with model scale, with larger models better equipped to leverage additional semantic information. 
Future work will focus on integrating this semantic understanding with autonomous driving planners, developing efficient interfaces that translate semantic insights into actionable information for path planning and decision-making algorithms.


{\small
\bibliographystyle{ieee_fullname}
\bibliography{main}

\begin{thebibliography}{10}\itemsep=-1pt

\bibitem{PaLMSL}
et~al. Aakanksha~Chowdhery.
\newblock Palm: Scaling language modeling with pathways.
\newblock {\em J. Mach. Learn. Res.}, 24:240:1--240:113, 2022.

\bibitem{AttentionIA}
et~al. Ashish~Vaswani.
\newblock Attention is all you need.
\newblock In {\em Neural Information Processing Systems}, 2017.

\bibitem{Bakr2023CoT3DRefCD}
Eslam~Mohamed Bakr, Mohamed Ayman, Mahmoud Ahmed, Habib Slim, and Mohamed Elhoseiny.
\newblock Cot3dref: Chain-of-thoughts data-efficient 3d visual grounding.
\newblock {\em ArXiv}, abs/2310.06214, 2023.

\bibitem{caron2021emerging}
Mathilde Caron, Hugo Touvron, Ishan Misra, Herv{\'e} J{\'e}gou, Julien Mairal, Piotr Bojanowski, and Armand Joulin.
\newblock Emerging properties in self-supervised vision transformers.
\newblock In {\em Proceedings of the IEEE/CVF International Conference on Computer Vision}, pages 9650--9660, 2021.

\bibitem{vil3drel}
Shizhe Chen, Pierre-Louis Guhur, Makarand Tapaswi, Cordelia Schmid, and Ivan Laptev.
\newblock Language conditioned spatial relation reasoning for 3d object grounding, 2022.

\bibitem{cordts2016cityscapesdatasetsemanticurban}
Marius Cordts, Mohamed Omran, Sebastian Ramos, Timo Rehfeld, Markus Enzweiler, Rodrigo Benenson, Uwe Franke, Stefan Roth, and Bernt Schiele.
\newblock The cityscapes dataset for semantic urban scene understanding, 2016.

\bibitem{unsloth}
Michael~Han Daniel~Han and Unsloth team.
\newblock Unsloth, 2023.

\bibitem{Ding2023HiLMDTH}
Xinpeng Ding, Jianhua Han, Hang Xu, Wei Zhang, and X. Li.
\newblock Hilm-d: Towards high-resolution understanding in multimodal large language models for autonomous driving.
\newblock {\em ArXiv}, abs/2309.05186, 2023.

\bibitem{Yin2023LAMMLM}
Zhen fei Yin, Jiong Wang, Jianjian Cao, Zhelun Shi, Dingning Liu, Mukai Li, Lu Sheng, Lei Bai, Xiaoshui Huang, Zhiyong Wang, Wanli Ouyang, and Jing Shao.
\newblock Lamm: Language-assisted multi-modal instruction-tuning dataset, framework, and benchmark.
\newblock {\em ArXiv}, abs/2306.06687, 2023.

\bibitem{lama3herdmodels}
Aaron Grattafiori and et al.
\newblock The llama 3 herd of models, 2024.

\bibitem{mamba}
Albert Gu and Tri Dao.
\newblock Mamba: Linear-time sequence modeling with selective state spaces, 2023.

\bibitem{Guo2023ViewReferGT}
Ziyu Guo, Yiwen Tang, Renrui Zhang, Dong Wang, Zhigang Wang, Bin Zhao, and Xuelong Li.
\newblock Viewrefer: Grasp the multi-view knowledge for 3d visual grounding.
\newblock {\em 2023 IEEE/CVF International Conference on Computer Vision (ICCV)}, pages 15326--15337, 2023.

\bibitem{lora}
Edward~J. Hu, Yelong Shen, Phillip Wallis, Zeyuan Allen-Zhu, Yuanzhi Li, Shean Wang, Lu Wang, and Weizhu Chen.
\newblock Lora: Low-rank adaptation of large language models, 2021.

\bibitem{3dsceneobjectlevel}
Haifeng Huang, Yilun Chen, Zehan Wang, Rongjie Huang, Runsen Xu, Tai Wang, Luping Liu, Xize Cheng, Yang Zhao, Jiangmiao Pang, and Zhou Zhao.
\newblock Chat-scene: Bridging 3d scene and large language models with object identifiers, 2024.

\bibitem{lama2}
et~al. Hugo~Touvron.
\newblock Llama 2: Open foundation and fine-tuned chat models.
\newblock {\em ArXiv}, abs/2307.09288, 2023.

\bibitem{jatavallabhula2023conceptfusionopensetmultimodal3d}
Krishna~Murthy Jatavallabhula, Alihusein Kuwajerwala, Qiao Gu, Mohd Omama, Tao Chen, Alaa Maalouf, Shuang Li, Ganesh Iyer, Soroush Saryazdi, Nikhil Keetha, Ayush Tewari, Joshua~B. Tenenbaum, Celso~Miguel de Melo, Madhava Krishna, Liam Paull, Florian Shkurti, and Antonio Torralba.
\newblock Conceptfusion: Open-set multimodal 3d mapping, 2023.

\bibitem{Jia2024SceneVerseS3}
Baoxiong Jia, Yixin Chen, Huangyue Yu, Yan Wang, Xuesong Niu, Tengyu Liu, Qing Li, and Siyuan Huang.
\newblock Sceneverse: Scaling 3d vision-language learning for grounded scene understanding.
\newblock In {\em European Conference on Computer Vision}, 2024.

\bibitem{Jin2023ADAPTAD}
Bu Jin, Xinyi Liu, Yupeng Zheng, Pengfei Li, Hao Zhao, T. Zhang, Yuhang Zheng, Guyue Zhou, and Jingjing Liu.
\newblock Adapt: Action-aware driving caption transformer.
\newblock {\em 2023 IEEE International Conference on Robotics and Automation (ICRA)}, pages 7554--7561, 2023.

\bibitem{kerbl20233dgaussiansplattingrealtime}
Bernhard Kerbl, Georgios Kopanas, Thomas Leimkühler, and George Drettakis.
\newblock 3d gaussian splatting for real-time radiance field rendering, 2023.

\bibitem{kerr2023lerflanguageembeddedradiance}
Justin Kerr, Chung~Min Kim, Ken Goldberg, Angjoo Kanazawa, and Matthew Tancik.
\newblock Lerf: Language embedded radiance fields, 2023.

\bibitem{opensceneunderstanding}
Sergey Linok, Tatiana Zemskova, Svetlana Ladanova, Roman Titkov, Dmitry Yudin, Maxim Monastyrny, and Aleksei Valenkov.
\newblock Beyond bare queries: Open-vocabulary object grounding with 3d scene graph, 2024.

\bibitem{malarz2024gaussiansplattingnerfbasedcolor}
Dawid Malarz, Weronika Smolak, Jacek Tabor, Sławomir Tadeja, and Przemysław Spurek.
\newblock Gaussian splatting with nerf-based color and opacity, 2024.

\bibitem{Malla2022DRAMAJR}
Srikanth Malla, Chiho Choi, Isht Dwivedi, Joonhyang Choi, and Jiachen Li.
\newblock Drama: Joint risk localization and captioning in driving.
\newblock {\em 2023 IEEE/CVF Winter Conference on Applications of Computer Vision (WACV)}, pages 1043--1052, 2022.

\bibitem{mildenhall2020nerfrepresentingscenesneural}
Ben Mildenhall, Pratul~P. Srinivasan, Matthew Tancik, Jonathan~T. Barron, Ravi Ramamoorthi, and Ren Ng.
\newblock Nerf: Representing scenes as neural radiance fields for view synthesis, 2020.

\bibitem{openai2024gpt35}
OpenAI.
\newblock Gpt-3.5 turbo api, 2024.
\newblock Accessed: 2024-07-29.

\bibitem{qin2024langsplat3dlanguagegaussian}
Minghan Qin, Wanhua Li, Jiawei Zhou, Haoqian Wang, and Hanspeter Pfister.
\newblock Langsplat: 3d language gaussian splatting, 2024.

\bibitem{radford2021learning}
Alec Radford, Jong~Wook Kim, Chris Hallacy, Aditya Ramesh, Gabriel Goh, Sandhini Agarwal, Girish Sastry, Amanda Askell, Pamela Mishkin, Jack Clark, et~al.
\newblock Learning transferable visual models from natural language supervision.
\newblock In {\em International conference on machine learning}, pages 8748--8763. PMLR, 2021.

\bibitem{shi2023le3dgaussians}
Jin-Chuan Shi, Miao Wang, Hao-Bin Duan, and Shao-Hua Guan.
\newblock Language embedded 3d gaussians for open-vocabulary scene understanding, 2023.

\bibitem{qwen2.5}
Qwen Team.
\newblock Qwen2.5: A party of foundation models, September 2024.

\bibitem{Tian2024TokenizeTW}
Ran Tian, Boyi Li, Xinshuo Weng, Yuxiao Chen, Edward Schmerling, Yue Wang, B. Ivanovic, and Marco Pavone.
\newblock Tokenize the world into object-level knowledge to address long-tail events in autonomous driving.
\newblock 2024.

\bibitem{GPT3}
et~al. Tom B.~Brown.
\newblock Language models are few-shot learners.
\newblock {\em ArXiv}, abs/2005.14165, 2020.

\bibitem{tschernezki2022neuralfeaturefusionfields}
Vadim Tschernezki, Iro Laina, Diane Larlus, and Andrea Vedaldi.
\newblock Neural feature fusion fields: 3d distillation of self-supervised 2d image representations, 2022.

\bibitem{oord2018neuraldiscreterepresentationlearning}
Aaron van~den Oord, Oriol Vinyals, and Koray Kavukcuoglu.
\newblock Neural discrete representation learning, 2018.

\bibitem{Wu2023LanguagePF}
Dongming Wu, Wencheng Han, Tiancai Wang, Yingfei Liu, Xiangyu Zhang, and Jianbing Shen.
\newblock Language prompt for autonomous driving.
\newblock {\em ArXiv}, abs/2309.04379, 2023.

\bibitem{xie2021segformer}
Enze Xie, Wenhai Wang, Zhiding Yu, Anima Anandkumar, Jose~M Alvarez, and Ping Luo.
\newblock Segformer: Simple and efficient design for semantic segmentation with transformers.
\newblock In {\em Neural Information Processing Systems (NeurIPS)}, 2021.

\bibitem{vlmgrounder}
Runsen Xu, Zhiwei Huang, Tai Wang, Yilun Chen, Jiangmiao Pang, and Dahua Lin.
\newblock Vlm-grounder: A vlm agent for zero-shot 3d visual grounding, 2024.

\bibitem{qwen2}
An Yang, Baosong Yang, Binyuan Hui, Bo Zheng, Bowen Yu, Chang Zhou, Chengpeng Li, Chengyuan Li, Dayiheng Liu, Fei Huang, et~al.
\newblock Qwen2 technical report.
\newblock {\em arXiv preprint arXiv:2407.10671}, 2024.

\bibitem{llmgrounder}
Jianing Yang, Xuweiyi Chen, Shengyi Qian, Nikhil Madaan, Madhavan Iyengar, David~F. Fouhey, and Joyce Chai.
\newblock Llm-grounder: Open-vocabulary 3d visual grounding with large language model as an agent, 2023.

\bibitem{LLM4DriveAS}
Zhenjie Yang, Xiaosong Jia, Hongyang Li, and Junchi Yan.
\newblock Llm4drive: A survey of large language models for autonomous driving.
\newblock {\em ArXiv}, abs/2311.01043, 2023.

\bibitem{Ye_2023}
Jianglong Ye, Naiyan Wang, and Xiaolong Wang.
\newblock Featurenerf: Learning generalizable nerfs by distilling foundation models.
\newblock In {\em 2023 IEEE/CVF International Conference on Computer Vision (ICCV)}. IEEE, Oct. 2023.

\bibitem{bagofthewords}
Mert Yuksekgonul, Federico Bianchi, Pratyusha Kalluri, Dan Jurafsky, and James Zou.
\newblock When and why vision-language models behave like bags-of-words, and what to do about it?, 2023.

\bibitem{zhang2024longclipunlockinglongtextcapability}
Beichen Zhang, Pan Zhang, Xiaoyi Dong, Yuhang Zang, and Jiaqi Wang.
\newblock Long-clip: Unlocking the long-text capability of clip, 2024.

\bibitem{zhi2021inplacescenelabellingunderstanding}
Shuaifeng Zhi, Tristan Laidlow, Stefan Leutenegger, and Andrew~J. Davison.
\newblock In-place scene labelling and understanding with implicit scene representation, 2021.

\bibitem{zhou2024feature3dgssupercharging3d}
Shijie Zhou, Haoran Chang, Sicheng Jiang, Zhiwen Fan, Zehao Zhu, Dejia Xu, Pradyumna Chari, Suya You, Zhangyang Wang, and Achuta Kadambi.
\newblock Feature 3dgs: Supercharging 3d gaussian splatting to enable distilled feature fields, 2024.

\bibitem{zhu2023minigpt4enhancingvisionlanguageunderstanding}
Deyao Zhu, Jun Chen, Xiaoqian Shen, Xiang Li, and Mohamed Elhoseiny.
\newblock Minigpt-4: Enhancing vision-language understanding with advanced large language models, 2023.

\bibitem{zürn2024wayvescenes101datasetbenchmarknovel}
Jannik Zürn, Paul Gladkov, Sofía Dudas, Fergal Cotter, Sofi Toteva, Jamie Shotton, Vasiliki Simaiaki, and Nikhil Mohan.
\newblock Wayvescenes101: A dataset and benchmark for novel view synthesis in autonomous driving, 2024.

\end{thebibliography}
}

\end{document}